\documentclass[sigconf,nonacm]{acmart}

\AtBeginDocument{%
  \providecommand\BibTeX{{%
    \normalfont B\kern-0.5em{\scshape i\kern-0.25em b}\kern-0.8em\TeX}}}

\usepackage[utf8]{inputenc}
\usepackage{pythonhighlight}
\usepackage{listings}
\usepackage{xcolor}
\usepackage{subfigure}
\usepackage{mdframed}

\definecolor{codegreen}{rgb}{0,0.6,0}
\definecolor{codegray}{rgb}{0.5,0.5,0.5}
\definecolor{codepurple}{rgb}{0.58,0,0.82}
\definecolor{backcolour}{rgb}{0.95,0.95,0.92}

\lstdefinestyle{mystyle}{
    commentstyle=\color{codegreen},
    keywordstyle=\color{magenta},
    numberstyle=\tiny\color{codegray},
    stringstyle=\color{codepurple},
    basicstyle=\ttfamily\footnotesize,
    breakatwhitespace=false,         
    breaklines=true,                 
    captionpos=b,                    
    keepspaces=true,                 
    numbers=left,                    
    numbersep=2pt,                  
    showspaces=false,                
    showstringspaces=false,
    showtabs=false,                  
    tabsize=1
}

\usepackage{amsmath,amsfonts,amsthm}
\usepackage{algorithmic}
\usepackage{multirow}
\usepackage[]{algorithm2e}
\begin{document}

\title{T-EMDE: Sketching-based global similarity for cross-modal retrieval}

\author{Barbara Rychalska}
\authornote{\label{equalContr}Both authors contributed equally to this research.}
\affiliation{%
  \institution{Synerise}
  \institution{Warsaw University of Technology}
  \country{Poland}
}

\author{Mikolaj Wieczorek\footnotemark[1]}
\affiliation{%
  \institution{Synerise}
  \institution{Warsaw University of Technology}
  \country{Poland}
}

\author{Jacek Dabrowski}
\affiliation{%
  \institution{Synerise}
  \country{Poland}
}

\renewcommand{\shortauthors}{B. Rychalska et al.}

\begin{abstract}
The key challenge in cross-modal retrieval is to find similarities between objects represented with different modalities, such as image and text. However, each modality embeddings stem from non-related feature spaces, which causes the notorious 'heterogeneity gap'. Currently, many cross-modal systems try to bridge the gap with self-attention. However, self-attention has been widely criticized for its quadratic complexity, which prevents many real-life applications. In response to this, we propose T-EMDE - a neural density estimator inspired by the recently introduced Efficient Manifold Density Estimator (EMDE) from the area of recommender systems. EMDE operates on \textit{sketches} - representations especially suitable for multimodal operations. However, EMDE is non-differentiable and ingests precomputed, static embeddings. With T-EMDE we introduce a trainable version of EMDE which allows full end-to-end training. In contrast to self-attention, the complexity of our solution is linear to the number of tokens/segments.  As such, T-EMDE is a drop-in replacement for the self-attention module, with beneficial influence on both speed and metric performance in cross-modal settings. It facilitates communication between modalities, as each global text/image representation is expressed with a standardized sketch histogram which represents the same manifold structures irrespective of the underlying modality. We evaluate T-EMDE by introducing it into two recent cross-modal SOTA models and achieving new state-of-the-art results on multiple datasets and decreasing model latency by up to 20\%.
\end{abstract}

\maketitle

\section{Introduction}
Cross-modal retrieval refers to expressing a single entity by means of various modalities, such as image, text or sound. For example, a garden party scene can be identified either by its photo, textual description (\textit{"a group of people laughing and talking"}), or the recorded sound of the scene. Image-text matching is particularly common, e.g. in database retrieval tasks of photos via their textual descriptions \cite{Li_2017_CVPR,pmlr-v69-deshpande17a,8682632}, image captioning \cite{10.1145/3295748,Yao_2018_ECCV,Huang_2019_ICCV}, and  multimodal  neural  machine  translation 
\cite{nishihara-etal-2020-supervised,10.1145/3372278.3390674}. 

Cross-modal matching remains a challenge due to the 'heterogeneity gap', caused by the fact that modalities are represented by inherently nonmatching feature spaces \cite{Wang2019a_Cross-modal_Scene_Graph_Matching, Li2019_Visual_Semantic_Reasoning_for_Image-Text_Matching, Diao2021_sgraf}. Cross-modal methods strive to bridge the gap with various neural architectures, such as Graph Neural Networks (GNNs) and multiple forms of attention \cite{Zhang_contextAware, Lee2018_scan, Diao2021_sgraf, Wang2019_Position_Focused_Attention, Chen2020_IMRAM} in order to compute per-modality vectors which can be compared. The similarity (or distance) between these vectors is used to compute the probability of a correct match between image and caption. Unfortunately, both GNNs and attention networks are known for poor scalability and specific performance issues \cite{ZENG2021166, pmlr-v80-chen18p,Bai2020RippleWT}. 

An important flavor of attention which is often found in cross-modal systems is the self-attention. It is used in various deep learning domains to obtain a summary representation of text or image. Self-attention has proven very successful, especially in the Transformer architecture \cite{Vaswani2017_attentionAllYouNeed}, which has has led to many performance breakthroughs in areas such as machine translation, language modeling, and image classification \cite{Su_VLBERT, Raffel2020_Exploring_the_Limits_of_Transfer, Wu2020_VisualTransformer, Dosovitskiy2020_image16words}. Yet, the core operation of self-attention is the dot product between a sequence representation and itself, which results in quadratic complexity. This characteristic makes self-attention inefficient in many real-life industrial applications. For this reason, attempts are made at devising more efficient architectures which could replace self-attention \cite{bello2021lambdanetworks,verma2021beyond}.

In this paper we address the above mentioned problems with an efficient neural module which can serve as a drop-in replacement for self-attention, which we call T-EMDE. Our proposed solution is especially appropriate for computation of global representations of text and image in multimodal retrieval scenarios. T-EMDE is inspired by the recently introduced EMDE (Efficient Manifold Density Estimator) \cite{emde} which has proven competitive in multimodal recommendation scenarios. We design  T-EMDE with three main goals in mind:
\begin{itemize}
    \item \textbf{High efficiency.} T-EMDE has linear complexity with respect to the sequence size (text tokens or image segments). This stands in contrast to quadratic complexity of self-attention.
    \item \textbf{Unified cross-modal representation.} T-EMDE maps any modality to the same type of representation - a semi-histogram on multidimensional data manifolds. This unified representation bridges the problem of heterogeneity gap.
    \item \textbf{Differentiability.} While the EMDE architecture has been shown to achieve high results and avoid scalability problems characteristic for complex neural architectures, it is a non-differentiable algorithm which prevents the fine-tuning of embeddings. With T-EMDE we exploit the same ideas which give EMDE its advantages, while making it fully trainable.
\end{itemize}

We evaluate T-EMDE by introducing it into two recent high-performing models which use self-attention: SAF (Similarity Attention Filtration) and SGR (Similarity Graph Reasoning) \cite{Diao2021_sgraf}.
By introducing T-EMDE, we are able to reach new state-of-the-art results on MSCOCO and Flickr30k datasets. We observe especially significant gains in the Recall@1 metric, representing the quality of the top returned recommendation. In multiple cases, the gains in Recall@1 are well over 1 pp. with comparison to the baseline SAF and SGR. At the same time, due to reduced complexity T-EMDE lowers model inference latency by up to 20\%. We propose T-EMDE as a step in the direction of replacing computationally expensive self-attention with more efficient approaches.

\section{Related Work}
\subsection{Text-image alignment}
A number of prior works \cite{Wang2015_Learning_Deep_Structure-Preserving, Faghri2017_VSEpp, Zheng2017_Dual-Path_Convolutional} focused on matching global image and text features in a joint embedding space. The aim was to learn to map both modalities to the same vector space, where the similarity measure could be applied for the cross-modal retrieval. One of the shortcomings of methods using only global representation was their ignorance of the fact that the notion of similarity may arise from an aggregation of numerous local similarities, such that salient objects in the image and keywords in the text. 
To address the problem, \cite{Karpathy2014} explored mutual relations between image regions and words a in text. The authors proposed to infer a similarity score between each pair of an image region and each word to define the alignment of the corresponding fragment features in the shared embedding space. This approach of computing similarity on the level of a smaller units from each modality was explored in many subsequent works.

Since the Transformer architecture proved successful in various domain, the self-attention mechanism has been explored by many researchers in cross-modal retrieval settings \cite{Song2019_Polysemous, Hu2019_Multi-level_visual-semantic, Chen2020_IMRAM, Weia_Multi-Modality_Cross_Attention, Zhang_contextAware, Lee2018_scan, Diao2021_sgraf}. 
\cite{Wang2019_Position_Focused_Attention} introduced Position Focused Attention Network (PFAN), which incorporates object position clues into the learning architecture. Images are split into blocks and attention mechanism is used to generate meaningful region features that are later fed into visual-textual attention to learn the joint-embedding space.
\cite{Chen2020a_Expressing_Objects_just_like_Words} proposed Dual Path Recurrent Neural Network (DP-RNN), which symmetrically processes images and sentences. It ‘reads’ the image objects in the order the text indicates, reorders them based on the object-word matching and an RNN learns joint information embeddings of semantically related objects. Finally, a module consisting of attention and self-attention mechanism is applied to compute the image-text similarity from recurrent visual and textual features.
\cite{Zhang_contextAware} observe that many attention-based methods ignore the fact that image segments or words might have various semantic meaning that depends on the global context, which in turn was defined as intra-modal and inter-modal relations. They devised Context-Aware Attention Network (CAAN), which aggregates context information of alignments between modalities and within a single modality simultaneously.
\cite{Chen2020_IMRAM} proposed to use Iterative Matching with Recurrent Attention Memory (IMRAM) method to capture the mutual image and text alignments iteratively. It consists of two main parts: (1) a cross-modal attention applied multiple times that builds the fine-grained correspondence between modalities progressively, (2) a memory distillation unit which aggregates alignment knowledge from earlier steps and passes it to later ones.

\cite{Li2019_Visual_Semantic_Reasoning_for_Image-Text_Matching} is an example of a Graph Convolutional Network model which generates visual features of salient objects and their semantic relationship. The gate and memory mechanism \cite{Chung2014_empiricalEvaluationGRU} is used to perform semantic reasoning, which considers both local and global semantic correlations. It enables the model to select information that contributes to a meaningful representation of the whole scene and ignore redundant cues at the same time. \cite{Wang2019a_Cross-modal_Scene_Graph_Matching} introduces Scene Graph Matching (SGM) - an approach using separate graphs for each modality to capture object features and their mutual relationships in corresponding modality. When the object-level and relationship-level features are extracted from both modalities, the final matching on both levels takes place.

\cite{Lee2018_scan} proposed Stacked Cross Attention Network (SCAN), which is trained to discover the full latent alignment between salient objects and words. The SCAN can be defined as two complimentary formulations, where each modality is used as a context to each other and two separate similarity scores are inferred. The alignment is done on the local-region level, which is next aggregated to build a global representation. While SCAN used the cross-modal attention with success it lacked the self-attention part. \cite{Diao2021_sgraf} builds upon SCAN, extends it with self-attention and applies it to each modality separately. It introduced two modules - Similarity Attention Filtration (SAF) and Similarity Graph Reasoning (SGR) to boost the accuracy of alignments between modalities.
SGR is based on a graph convolutional neural network (GCNN). It builds a similarity graph that propagates similarities at local and global levels to obtain the alignment scores.
Similarity Attention Filtration (SAF) module was introduced as the authors noticed that using the local alignments generally boosts the performance, but incorporating the less relevant elements into the global representation decreases the final results. SAF aim is to decide which alignments are significant and which should be filtered out. The SGRAF model is an ensemble of SRG and SAF models, which similarity scores are averaged before the retrieval takes place.

\subsection{Attention Mechanism}
Attention is a mechanism that allows the network to learn correspondence between input and target sequence/set elements.
One of the first works that introduced the popular attention mechanism was \cite{Bahdanau2015_Neural_machine_translation_by_jointly}. The authors proposed a simple additive module that computed 'attention scores' for each input element. The weights of the alignments were computed by feeding RNN hidden state and a word latent vector into a single-layer feedforward neural network. This allowed the RNN decoder to attend to any part of the  source sentence and in turn it increased the quality of machine translation task.
A similar approach can be found in Computer Vision, where attention was used to help the network find relevant images patches and object in the image to generate captions \cite{Xu2015_Show_attend_and_tell}.

The Transformer architecture  \cite{Vaswani2017_attentionAllYouNeed} introduced a scaled dot product self-attention computed over three elements: the Query, Key, and Value. This can be deemed as an extension of the attention from \cite{Bahdanau2015_Neural_machine_translation_by_jointly}. The Transformer architecture computes the self-attention multiple times separately, which is known as multi-head attention. According to the authors this allows the model to attend to information from different representation subspaces at different positions. Furthermore, the attention mechanism was extended by residual connections and normalization layers. Transformer model and its self-attention mechanism has started to dominate various domains. For example, it achieved state-of-the-art results in machine translation \cite{Tay, Raffel2020_Exploring_the_Limits_of_Transfer},  text summarizing \cite{Raffel2020_Exploring_the_Limits_of_Transfer}, visual question answering \cite{Su_VLBERT, Tan_LXMERT}.

However, the Transformer architecture is notorious for its difficulty to train and its demand for large amounts of data. Most current state-of-the-art solutions in cross-modal retrieval are non-Transformer models, which nevertheless use simplified dot-product self-attention modules commonly \cite{Diao2021_sgraf,DBLP:conf/ijcai/XuLYDL19, Song2019_Polysemous,9292444,DBLP:conf/ijcai/XuLYDL19, Wang2019_Position_Focused_Attention, Hu2019_Multi-level_visual-semantic,Chen2020_IMRAM,Lee2018_scan}.

\subsection{Optimizing Self-Attention}
Unfortunately, the time and memory complexity of self-attention is quadratic as it needs to compute the weights between each pair of inputs. It prevents efficient scalability of attention-based models, especially where longer sequences are present in the data.
To tackle this problem many researchers introduced the extensions and improvements, which are however primarily dedicated for the Transformer architecture.

Some works attempt to limit Transformer self-attention by attending only to some inputs according to a fixed block/patterns \cite{Beltagy_LongFormer,Dai_transformerXL, Qiu2019_BlockBERT,Tay2020_Sinkhorn, Child_sparseTransfomer}.
Another approach to improve the complexity of Transformer self-attention is to use low-rank factorization methods. \cite{Wang2020a_LinFormer, Tay} assumed low-rank structure of self-attention. They project the Transformer Key and Value vectors into a lower-dimensional representation.
Some works \cite{Choromanski_PerFormer, Katharopoulos2020_TransformerAreRNNs} used kernel functions to bring down the dimensionality of the elements on which the Transformer self-attention operates.
\cite{Wu2021_CentroidTransformer, Roy2020_RoutingTransformer} introduced centroids/clusters into the Transformer self-attention. The idea is to calculate the attention scores only among the tokens belonging to the same cluster/centroid. \cite{Kitaev_ReFormer} proposed Transformer LSH Attention, which uses Locally-Sensitive Hashing method to obtain hashes for both the Keys and the Queries.

Thus, most attempts at limiting the complexity of self-attention are dedicated for the Transformer's formulation of self-attention, assuming the creation of multiple Query, Key, Value vectors to build multi-head attention within a strictly defined neural architecture. Yet, few attempts are made at optimization or replacement of the basic formulation of (self) attention. This comes at a disadvantage to models in which the simple classic attention composed of a single dot product and weighted summation of vectors is the preferred choice \cite{Diao2021_sgraf, DBLP:conf/ijcai/XuLYDL19, Wang2019_Position_Focused_Attention, Hu2019_Multi-level_visual-semantic}. With T-EMDE, we mean to bridge the gap by introducing a conceptually simple and fast alternative to classic self-attention. Additionally, we aim at easy communication between various modality representations, which is not handled inherently by neither formulation of attention.

\begin{figure}
\includegraphics[width=\columnwidth]{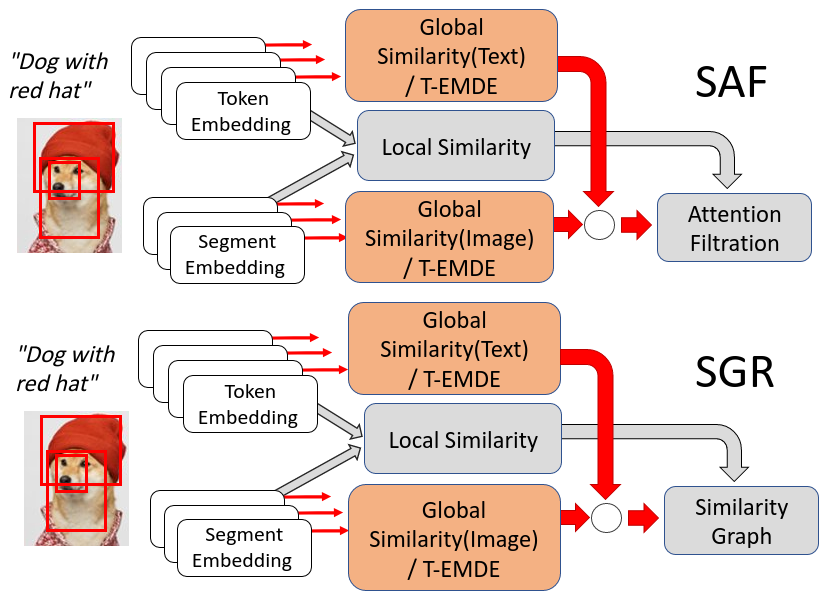}
\caption{An overview of the SAF and SGR architectures, showing where T-EMDE is introduced.} \label{fig:saf-sgr}
\end{figure}

\section{Proposed method}
In this section we describe the following architectures: EMDE \cite{emde} which serves as an inspiration model, T-EMDE - our proposed module, and SAF/SGR \cite{Diao2021_sgraf} - models which serve as scaffolds into which T-EMDE is introduced.

\subsection{Non-Trainable Efficient Manifold Density Estimator}
\begin{figure*}
\includegraphics[width=\textwidth]{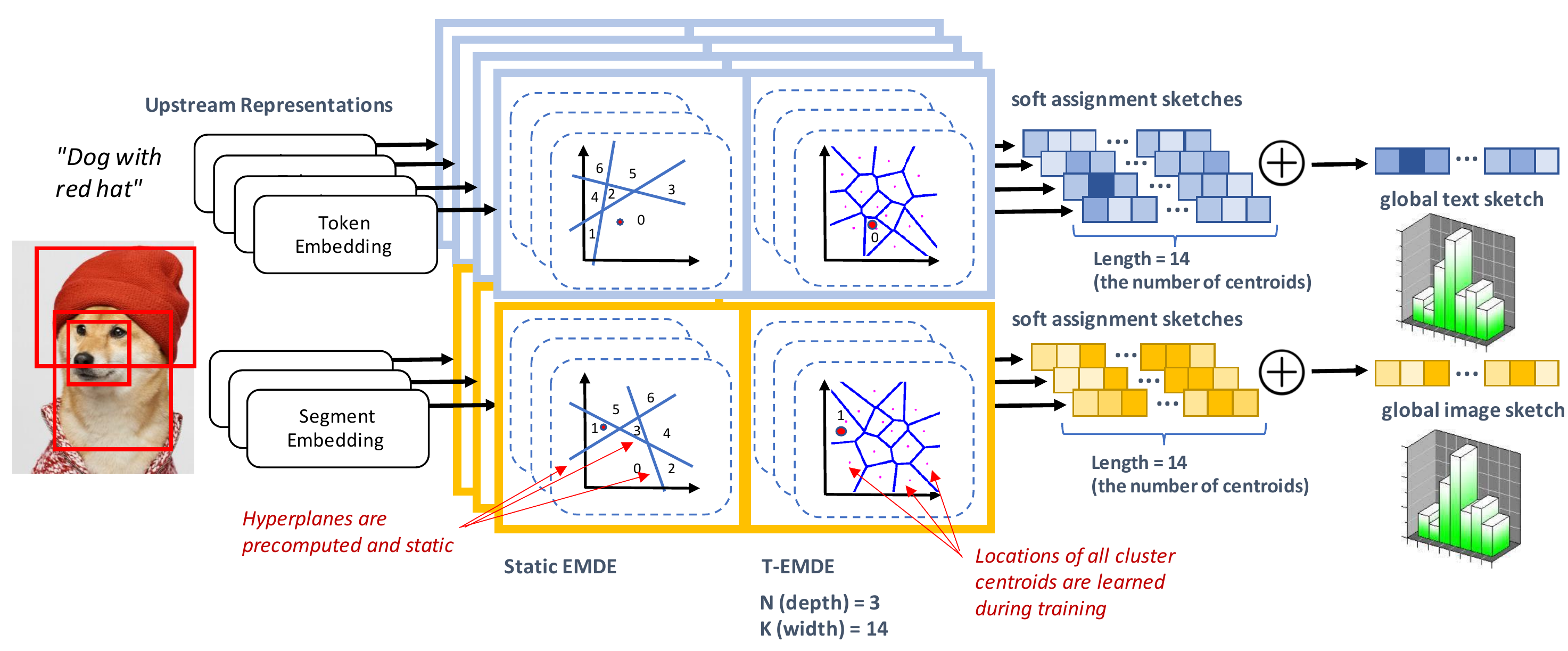}
\caption{T-EMDE module, with comparison to the static EMDE.} \label{fig:t-emde}
\end{figure*}

Efficient Manifold Density Estimator (EMDE) introduced in \cite{emde} is a probability density estimator inspired by Count-Min Sketch algorithm (CMS) and local sensitive hashing (LSH). The overview of the algorithm is shown in Figure \ref{fig:t-emde}. EMDE ingests input data represented by vectors embedded on manifolds spanned by various upstream representation learning methods. The manifolds are then partitioned via a data-dependent LSH method (DLSH). The partitioning method divides the manifold into regions, analogous to CMS \textit{buckets}. The purpose of manifold partitioning follows the logic of LSH: similar data points should be mapped to the same region. While a single region is large (typically 64-256 regions form a single partitioning covering the whole manifold, as described in \cite{emde}), multiple independent partitionings allow to obtain a high resolution map of the manifold via intersection or ensembling. 

The resulting region assignment vectors (\textit{sketches}) can be thought of as a form of a histogram. Each position within a sketch corresponds to a region, and each value represents the number of data points in the given region. For example, a sketch of a single item can take the form of $I = [0,0,1,0,0]$, which means that the data point in question is located in region number 2 out of 5 total regions. All items are represented with sketches of the same dimensionality, and representations of item sets are computed by simple summation of individual item sketches. For example, a shopping basket of 4 items can be represented with as sketch $S=[1,0,2,0,1]$ containing 4 items in total, one located in region number 0, two in region number 2, and one in region number 4. EMDE precomputes sketch representations of all items within the inventory, computes aggregate shopping basket sketches, and uses simple feed-forward network for mapping a basket sketch to a predicted sketch of items which will be likely bought next.

The algorithm is shown to achieve competitive results in product recommendation settings, which are often multimodal. Products can be characterized by user interactions (e.g. assignment to shopping baskets), textual descriptions, photos, and categorical features. In EMDE, all per-modality representations are transformed to the histogram-like sketches, which follow the same logic irrespective of the underlying modality or representation method. EMDE is shown to be fast and efficient - thanks to the application of the hashing-based CMS structure, it works in sublinear space. The neural architecture of EMDE is very simple, with just feed-forward layers.

However, in \cite{emde} the DLSH method is neither differentiable nor end-to-end trainable, due to the highly discontinuous binary-unary conversion being a key operation during assignment of inputs to region indices. Thus, manifold partitionings must be precomputed before actual training and are based on pretrained vectors, which prevents the model from learning own embedding vectors or fine-tuning embeddings derived from upstream representations. A drop-in replacement module for attention must be trainable, as otherwise it would block off the gradient in a whole portion of the underlying network. In particular, cross-modal architectures often do not use precomputed embeddings but rather train their own representations from scratch, which makes it obligatory for all parts of the network to be differentiable.

\subsection{Trainable Efficient Manifold Density Estimator}\label{temde}
Taking into the account all key characteristics which an attention replacement module should possess, we introduce the T-EMDE as a trainable version of EMDE. We retain the core properties of EMDE: high scalability due to the application of a hashing-based data structure and a unified representation used for all modalities to bridge the heterogeneity gap. Moreover, we aim to retain the ability of EMDE to model item sets, which in our case will be sets of tokens and image segments. The ability of computing a single object representation out of its composing part sequence (of variable size) makes T-EMDE a valid candidate for replacing self-attention.

T-EMDE aims to partition the underlying data manifold, similarly as EMDE. However, instead of a static assignment of inputs to specific regions of the manifold, we propose to use trainable centroids. The centroids are created within a space of arbitrary dimensionality and represent the basis of item representation - each token/image segment will be characterized by a vector of distances to all centroids. An overview of EMDE is shown in Figure \ref{fig:saf-sgr}, presenting its application to multiple modalities.

Our algorithm proceeds as follows:

\textbf{1. Centroid initialization.} We initialize a trainable centroid tensor $\mathbf{C}$ of dimensionality $N \times K \times D$. The tensor will hold $K$ centroids for each of $N$ independent manifold divisions. Independent manifold divisions are introduced with regard to the observation from \cite{emde} that a single manifold partitioning will usually not be perfect and multiple independent divisions can achieve an ensembling effect, which is beneficial to performance. Each centroid will be located within $D$-dimensional space and its coordinates will change during training to represent data assignments most appropriately. The parameter $N$ corresponds to sketch $depth$ from \cite{emde} (number of independent manifold partitionings), while $K$ corresponds to sketch $width$ (number of regions produced by each partitioning).

\textbf{2. Embedding projection.} Subsequently, we feed each item modality embedding to a simple linear layer to obtain a projected representation $\mathbf{X}$ of dimensionality $N \times K \times D$. This way we bring the input embedding into the $D$-dimensional space in which the cluster centroids live. After this transformation, input data representation can be represented by its proximity to centroids. The vectors $\mathbf{X}$ are batch-normalized.

\textbf{Distance computation.} After bringing input data into the centroid space, we compute the distance between each $\mathbf{X}$ and each centroid as $d = (\mathbf{X}-\mathbf{C})^2$. This represents a squared distance, but other distance metrics can be used if deemed appropriate. We then sum across $D$, to obtain squared euclidean distance vectors $\mathbf{X\_euclid}$ of dimensionality $N \times K$. Finally, the obtained representations are normalized with $\mathbf{Y} = softmax(\mathbf{X\_euclid})$ across the $K$ dimension. The resulting vectors can be thought of as representing soft assignments to $K$ centroids each, over $N$ independent space divisions.

T-EMDE can be implemented in a few lines of code. We attach its pseudocode written in PyTorch convention in Code Listing 1.

\begin{mdframed}[backgroundcolor=backcolour]
\begin{lstlisting}[caption=T-EMDE code, language=Python]
#given depth and width parameters
N = 20
K = 8
inner_dim = 8
#input representation size
input_dim = 1024

class Coder(nn.Module):
    def __init__(self):
        self.l1 = nn.Linear(input_dim, inner_dim * N * K)        
        self.bn1 = nn.BatchNorm1d(inner_dim * N * K)
        self.centroids = nn.Parameter(torch.randn(inner_dim * N * K))
        self.temp = nn.Parameter(torch.ones(1,))

    #encode individual tokens/image segments    
    def forward(self, x):
        x = self.l1(x)
        x = self.bn1(x)
        
        dists = (x-self.centroids)**2
        dists = dists.view(-1, N * K, inner_dim).sum(-1).view(-1, N, K)
        soft_assignments = F.softmax(-dists * self.temp, -1)
        soft_assignments = soft_assignments.view(-1, N * K)
        return soft_assignments 

class T_EMDE(nn.Module):
    def __init__(self):
        self.coder = Coder()
    
    #encode a whole text/image
    def forward(token_embeddings):
        encoded = self.coder(token_embeddings)
        summed = torch.sum(encoded, 0)
        normalized = l2norm(summed.view(-1, N, K), dim=-1).view(-1, K * N)
        return normalized
\end{lstlisting}
\end{mdframed}

It can be seen that T-EMDE exploits the feature space of various modalities, yet at the same time the output vector (the sketch) will represent solely the distances to centroids, losing all modality-specific characteristics. Comparison of two sketches coming from two T-EMDE coders for various modalities (e.g. text and image) will consist in defining the relationships between two sets of centroids.

\subsection{SAF and SGR}

SAF (Similarity Attention Filtration) and SGR (Similarity Graph Reasoning) are two recent state-of-the art models proposed by \cite{Diao2021_sgraf}. Like most recent cross-modal models, they are complex architectures composed of multiple submodules. They form the backbone of our proposed solution. Below we describe the composing modules of each solution (some are shared between models). An overview of SAF and SGR architectures is given in Figure \ref{fig:saf-sgr}.

\textbf{Image Representation.} Both models represent images as extracted segment embeddings. Such visual features are computed following \cite{DBLP:conf/cvpr/00010BT0GZ18} to extract K region-level visual features, with the Faster R-CNN \cite{NIPS2015_14bfa6bb} model pretrained on Visual Genomes \cite{DBLP:journals/ijcv/KrishnaZGJHKCKL17}. Only segment embeddings are considered, without recognizing additional features such as segment frame locations or captions. Each segment is projected through a fully-connected layer to transform it to a desired size.

\textbf{Text Representation.} Captions are tokenized with the \texttt{nltk} package \cite{Loper02nltk:the} and each token is represented by a randomly initialized embedding. The embeddings are fed to a bi-directional recurrent GRU network \cite{cho-etal-2014-properties} and token representations are obtained by averaging the hidden states from the forward and backward GRU pass.

\textbf{Global Similarity Module.} Global representations are computed for text and image separately with self-attention modules. First, all representations are run through separate sequences of multiple linear layers alternating with batch normalization, to form \textit{local} $\mathbf{L}$ and \textit{global} $\mathbf{G}$ variants of each modality representations. \textit{Global} representation is obtained by averaging per-segment or per-token representations, while \textit{local} representation consists of token/segment-level vectors. Then, self-attention is computed with a dot product over the two inputs, with $\mathbf{G}$ repeated to match the number of segments/tokens within $\mathbf{L}$:
\[
attn\_weights = softmax(\mathbf{L} \times repeat(\mathbf{G}, len(\mathbf{L}))) 
\]
\[
\mathbf{new\_G} = \sum_{j=1}^{n\_columns} (attn\_weights \cdot \mathbf{L})_{ij}
\]

After additional L2-normalization, $\mathbf{new\_G}$ is the final per-modality global representation. Global similarity vector is computed by subtracting the global caption  representation from the global image representation.

The Global Simialrity Module is one of the computationally heavy parts of the architecture, especially that it needs to be computed two times - always both for text and image. It is replaced with T-EMDE in our solution.

\textbf{Local Similarity Module.} Local Similarity Module also computes attention, but in contrast to Global Similarity Module it is done between the textual and image representations. It uses the cross-modal attention formula from \cite{Lee2018_scan} and obtains a summary textual-visual representation.

\textbf{Similarity Graph Module.} This module uses a Graph Neural Network architecture from \cite{DBLP:conf/iccv/Kuang0LLCLZ19} to model the strength of connection between tokens and image parts. Token/segment local representations together with the global representation computed by the Global Similarity Module are treated as graph nodes within a fully-connected graph. Edges are computed by weighted multiplication of input and output node representations, with each node representation  multiplied by trainable matrices $\mathbf{W}_{in}$ and $\mathbf{W}_{out}$. Similarity reasoning is performed by summing the edge representations of each node $p$, multiplied by the representations of neighbors of $p$, and running them through an extra layer with multiplication against a trainable matrix $\mathbf{W}_{r}$ with a nonlinearity. The global representation nodes are finally fed to a linear layer to return the text-image similarity score.

\textbf{Similarity Attention Filtration.}
Similarity Attention Filtration is designed to suppress tokens or segments which have a small influence on the final similarity (such as the function words "a", "be", etc.). It ingests vector similarity representations from previous modules (e.g. Global Similarity and Local Similarity) and runs them through additional linear layers, applying weights assessing the validity of each similarity alignment. A final linear layer ingesting weighted similarity representations returns the text-image similarity score.

As shown in Figure \ref{fig:saf-sgr}, SAF is composed of two Global Similarity Modules (each per modality), a Local Similarity Module, and the Attention Filtration which computes the final similarity scores. SGR also has two Global Similarity Modules and a Local Similarity Module, followed by Similarity Graph Module.

\subsubsection{T-EMDE in SAF/SGR}
T-EMDE can be easily introduced in place of the Global Similarity modules (or in fact any other formulation of self-attention). Thanks to the application of two very different neural architectures in SAF and SGR (a graph neural network and an attention filter), the versatility of application of T-EMDE to various architectures can be tested.

The module is applied in the same way to both SAF and SGR, replacing the Global Similarity modules for both image and text. In total, two T-EMDE modules are created, separately for image and text. The segment/token representations fed by the Image Representation and Token Representation modules are fed as input to respective T-EMDE modules. Individual cluster assignments of each token/segment are then retrieved. For each image and caption, the appropriate token/segment soft assignment vectors are summed, exploiting the additivity property of EMDE and Count-Min Sketch \cite{emde}. In EMDE, summation is done on item sketches which belong to one shopping basket/user in order to get a global basket/user representation. Similarly in cross-modal retrieval, summation can be thought of as aggregating multiple partwise histograms to get an overview profile of text/image - its global representation. The summation allows for keeping the global representation size constant irrespective of how many composing parts (tokens/segments) need to be aggregated. This mechanism allows us to handle the problem of variable sequence length and escape the necessity of comparing each item to all others which results in quadratic complexity of self-attention.

In contrast to Global Similarity module, the global representations are not subtracted, as corresponding centroids can get assigned to different places in the output sketches. Instead, we concatenate the text and image sketches (this can be done because their size is kept constant at all times) and map them to a smaller size with a linear layer, followed by ReLU activation.

\setlength{\tabcolsep}{3pt}
\begin{table}
\centering
\caption{Performance results on the MSCOCO 5K dataset (averaged over 5 runs).}\label{tab:5k-results}
\begin{tabular}{l|lll|lll}
\hline
 & \multicolumn{6}{c}{\textbf{MSCOCO 5K dataset}} \\
Model & \multicolumn{3}{c}{Text Retrieval} & \multicolumn{3}{c}{Image Retrieval}
 \\
 & R@1 & R@10 & MRR & R@1 & R@10 & MRR \\\hline
 
 SGM \cite{Wang2019a_Cross-modal_Scene_Graph_Matching} & 50.0 & 87.9 & - & 35.3 & 76.5 & - \\
 SCAN \cite{Lee2018_scan} & 50.4 & 90.0 & - & 38.6 & 80.4 & - \\
 CAAN \cite{Zhang_contextAware} & 52.5 & 90.9 & - & 41.2 &\textbf{82.9} & - \\
 VSRN \cite{Li2019_Visual_Semantic_Reasoning_for_Image-Text_Matching} & 53.0 & 89.4 & - & 40.5 & 81.1 & - \\
 IMRAM \cite{Chen2020_IMRAM} & 53.7 & 91.0 & - & 39.7 & 79.8 & - \\
  
  \hline\hline
  SAF noglobal & 53.7 & 90.4 & 0.663 & 39.8 & 80.0 & 0.532 \\
  SAF reported & 53.3 & 90.1 & - & 39.8 & 80.2 & - \\
  SAF reproduced & 55.6 & 90.7 & 0.677 & 40.3 & 80.3 & 0.536 \\
  
  T-EMDE & 56.7 & 90.7 & 0.682 & 40.3 & 80.4 & 0.537 \\ \hline
  SGR noglobal & 54.2 & 90.3 & 0.668 & 39.3 & 79.3 & 0.525  \\
  SGR reported & 56.9 & 90.5 & - & 40.2 & 79.8 & - \\
  SGR reproduced & 55.7 & 90.3 & 0.677 & 39.8 & 79.9 & 0.531 \\
  T-EMDE & 57.0 & 91.0 & 0.685 & 40.0 & 80.1 & 0.533 \\\hline
  SGRAF reported & 58.8  & 91.6  & - &	41.6 & 81.5 & -  \\
  SGRAF reproduced & 57.9 & 91.6 & 0.697 & \textbf{41.8} & 81.5 & 0.550  \\
  T-EMDE & \textbf{59.1} & \textbf{91.8} & \textbf{0.703} & \textbf{41.8} & 81.7 & \textbf{0.551} \\
\hline
\end{tabular}
\end{table}

\setlength{\tabcolsep}{3pt}
\begin{table*}
\centering
\caption{Performance results on Flickr30K and MSCOCO 1K datasets (averaged over 5 runs).}\label{tab1}
\label{tab:1k-results}
\begin{tabular}{l|llll|llll|llll|llll}
\hline
 & \multicolumn{8}{c}{\textbf{Flickr30K dataset}} & \multicolumn{8}{c}{\textbf{MSCOCO 1K dataset}}\\
Model & \multicolumn{4}{c}{Text Retrieval} & \multicolumn{4}{c}{Image Retrieval} & \multicolumn{4}{c}{Text Retrieval} & \multicolumn{4}{c}{Image Retrieval}
 \\
 & R@1 & R@5 & R@10 & MRR & R@1 & R@5 & R@10 & MRR  & R@1 & R@5 & R@10 & MRR & R@1 & R@5 & R@10 & MRR \\\hline
 
 CAAN \cite{Zhang_contextAware} & 70.1 & 91.6 & 97.2 & - & 52.8 & 79.0 & 87.9 & - & 75.5 & 95.4 & 98.5 & - & 61.3 & 89.7 & 95.2 & - \\
 DP-RNN \cite{Chen2020a_Expressing_Objects_just_like_Words} & 70.2 & 91.6 & 95.8 &  - & 55.5 & 81.3 & 88.2 & - & 75.3 & 95.8 & 98.6 & - & 62.5 & 89.7 & 95.1 & - \\
 PFAN \cite{Wang2019_Position_Focused_Attention} & 70.0 & 91.8 & 95.0 &  - & 50.4 & 78.7 & 86.1 & - & 76.5 & \textbf{96.3} & \textbf{99.0} & - & 61.6 & 89.6 & 95.2 & - \\
 VSRN \cite{Li2019_Visual_Semantic_Reasoning_for_Image-Text_Matching} & 71.3 & 90.6 & 96.0 & - & 54.7 & 81.8 & 88.2 & - & 76.2 & 94.8 & 98.2 & - & 62.8 & 89.7 & 95.1 & - \\
 IMRAM \cite{Chen2020_IMRAM} & 74.1 & 93.0 & 96.6 & - & 53.9 & 79.4 & 87.2 & - & 76.7 & 95.6 & 98.5 & - & 61.7 & 89.1 & 95.0 & - \\
  
  \hline\hline
  SAF no global & 71.1 & 91.6 & 96.3 & 0.803 & 54.1 & 80.7 & 87.3 & 0.660 & 76.2 & 95.5 & 98.1 & 0.845 & 61.5 & 89.3 & 95.1 & 0.736 \\
  SAF reported & 73.7 & 93.3 & 96.3 & - & 56.1 & 81.5 & 88.0 & - & 76.1 & 95.4 & 98.3 & - & 61.8 & 89.4 & 95.3 & - \\
  SAF reproduced & 74.9 & 93.9 & 97.1 & 0.833&
  56.3 & 81.8 & 88.0 & 0.676 & 77.2 & 95.7 & 98.5 & 0.853 & 62.1 & 89.7 & 95.4 & 0.741 \\
  
  T-EMDE & 75.2 & 94.2 & 97.1 & 0.829 &
  57.1 & 82.2 & 88.3 & 0.682 & 78.3 & 95.7 & 98.5 & 0.858 & 62.3 & 89.7 & 95.2 & 0.745 \\\hline
  SGR no global & 52.8 & 83.7 & 92.8 & 0.664 &  49.6 & 77.3 & 85.1 & 0.618 & 76.6 & 95.8 & 98.4 & 0.848 &  60.9 & 89.0 & 95.1 & 0.731 \\
  SGR reported & 75.2 & 93.3 & 96.6 & - & 56.2 & 81.0 & 86.5 & - & 78.0 & 95.8 & 98.2 & - & 61.4 &  89.3 & 95.4 & -  \\
  SGR reproduced & 76.3 & 93.2 & 96.9 & 0.835 & 55.2 & 80.7 & 88.0 & 0.666 & 76.9 & 95.5 & 98.5 & 0.849 & 61.5 & 89.4 & 95.4 & 0.737 \\
  T-EMDE & 77.5 & 93.1 & 97.2 & 0.845 & 56.9 & 82.0 & 87.5 & 0.679 & 77.1 & 95.9 & 98.5 & 0.852 & 61.6 & 89.5 & 95.1 & 0.737 \\\hline
  SGRAF reported & 78.4	& \textbf{94.6} & \textbf{97.5} & - & 58.2 & 83.0	& 89.1 & - & 79.2 & 96.2 & 98.5 & - & 63.2 & \textbf{90.7} & 96.1 & -  \\
  SGRAF reproduced & 77.5 & 94.4 & 97.0 & 0.849 & 58.4 & 83.2 & 89.0 &0.693 & 78.9 & 96.2 & 98.9 & 0.864 & \textbf{63.6} & 90.4 & \textbf{95.8} & \textbf{0.752} \\
  T-EMDE & \textbf{78.8} & 94.4 & \textbf{97.5} & \textbf{0.858} & \textbf{59.6} & \textbf{83.6} & \textbf{89.2} & \textbf{0.702} & \textbf{79.6} & \textbf{96.3} & 98.7 & \textbf{0.868} & 63.5 & 90.4 & 95.6 & \textbf{0.752} \\\hline
\end{tabular}
\end{table*}

\section{Experiments}

\subsection{Datasets}
Our evaluation is aligned with \cite{Diao2021_sgraf} for fair comparison, we reuse their evaluation code\footnote{\url{https://github.com/Paranioar/SGRAF}} for full credibility and evaluate on the same datasets. Thus, we use two popular cross-modal datasets: MSCOCO \cite{DBLP:conf/eccv/LinMBHPRDZ14} and Flickr30K \cite{DBLP:journals/tacl/YoungLHH14} datasets. The MSCOCO dataset s a large-scale object detection, segmentation, and captioning dataset. It contains 123,287 photos, mainly from daily life scenes. Each image is annotated with 5 captions produced by crowdworkers from  Amazon  Mechanical Turk. We use a popular cross-modal retrieval split  (the “Karpathy” split) which assigns 113,287 images for training, 5,000 images with 25,000 matching captions for validation and 5,000 images with 25,000 matching captions for testing. We use two evaluation protocols:
\begin{itemize}
\item MSCOCO 1K - the results are computed by averaging over 5 folds, each of 1000 test images.
\item MSCOCO 5K - the results are computed on the full set of 5000 images. 
\end{itemize}
The Flickr30K dataset contains 31,783 images with 5 corresponding captions each. Generally, the Flickr30k captions are much longer and are in many cases more detailed than in MSCOCO. \cite{coco_flickr_comparison}. Following the split in \cite{NIPS2013_7cce53cf}, we use 1,000 images for validation, 1,000 images for testing and the rest for training.

\subsection{Performance Metrics}
As our main performance measure we adapt the Recall@k (R@k) metric applied in \cite{Diao2021_sgraf}. Recall@k can be understood as the proportion of relevant items found in the top-\textit{k} list of retrieved items. 
Since the performance of recent cross-modal retrieval models seems to be saturating and the results are often very similar, we also employ an additional performance measure: the mean reciprocal rank (MRR), a popular metric in information retrieval. MRR is the average of the reciprocals of the ranks of the relevant items from the whole query set $Q$:

        $$ MRR = \frac{1}{|Q|} \sum_{i=1}^{|Q|}{\frac{1}{rank_i}}. $$  
        
We compute MRR on T-EMDE, SAF and SGR in order to differentiate their performance results better.

\subsection{Training configuration}

For the MSCOCO dataset we set $N=20$ and $K=8$, and for the Flickr30K - $N=16$ and $K=8$, selected experimentally. The inner dimension of the centroid space $D$ is always set to 8 and the centroid distance metric is euclidean distance. All other parameters are as in \cite{Diao2021_sgraf}. T-EMDE/SAF and T-EMDE/SGR are trained separately, and T-EMDE/SGRAF results are computed as average of similarities returned by SAF and SGR (as is done in \cite{Diao2021_sgraf}).

\begin{figure*}
\subfigure["boy"]{%
  \includegraphics[width=0.15\textwidth]{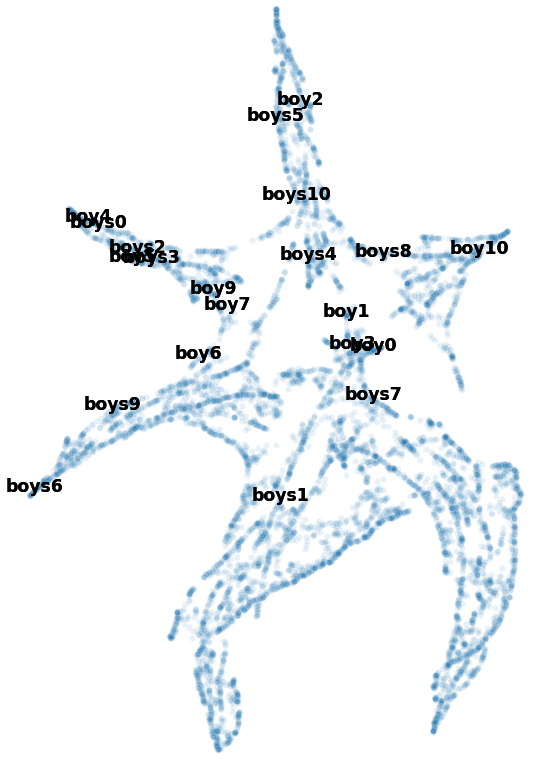}%
}
\subfigure["tree"]{%
  \includegraphics[width=0.15\textwidth]{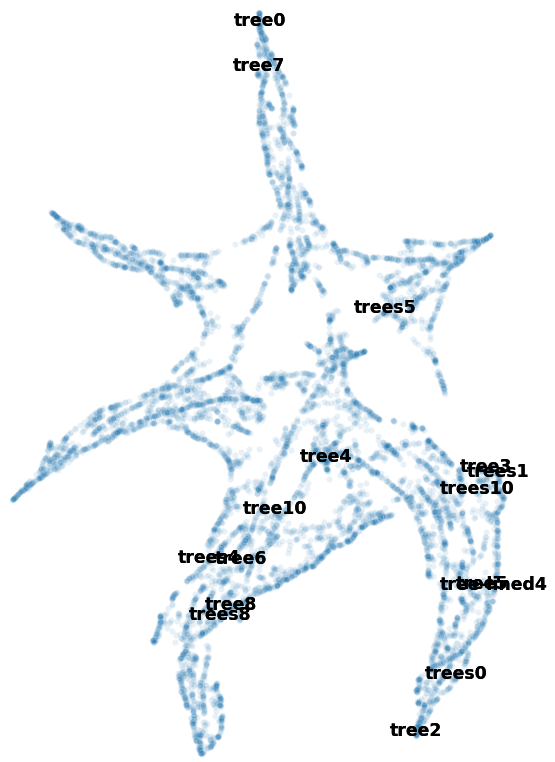}%
} 
\subfigure["there"]{%
  \includegraphics[width=0.15\textwidth]{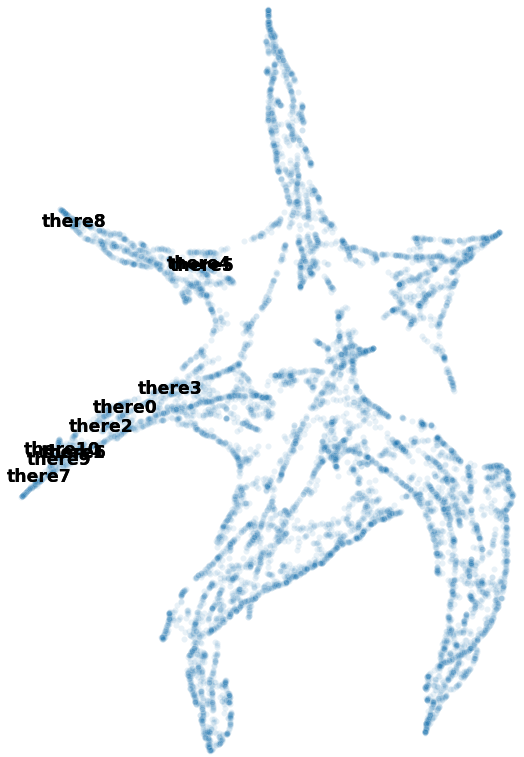}%
} 
\subfigure["these"]{%
  \includegraphics[width=0.15\textwidth]{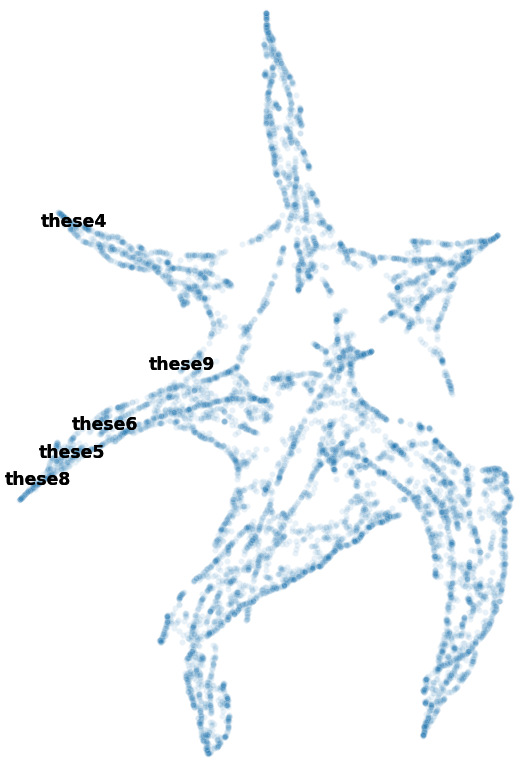}%
} 
\subfigure["big"]{%
  \includegraphics[width=0.15\textwidth]{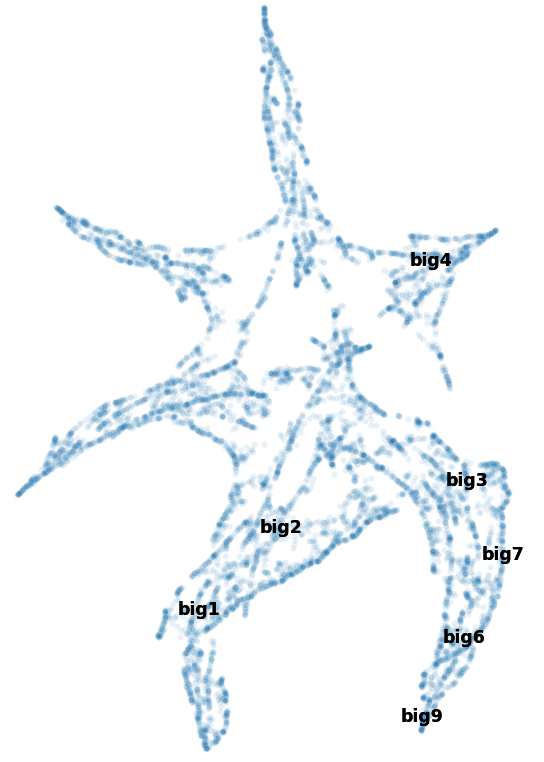}%
} 
\subfigure["small"]{%
  \includegraphics[width=0.15\textwidth]{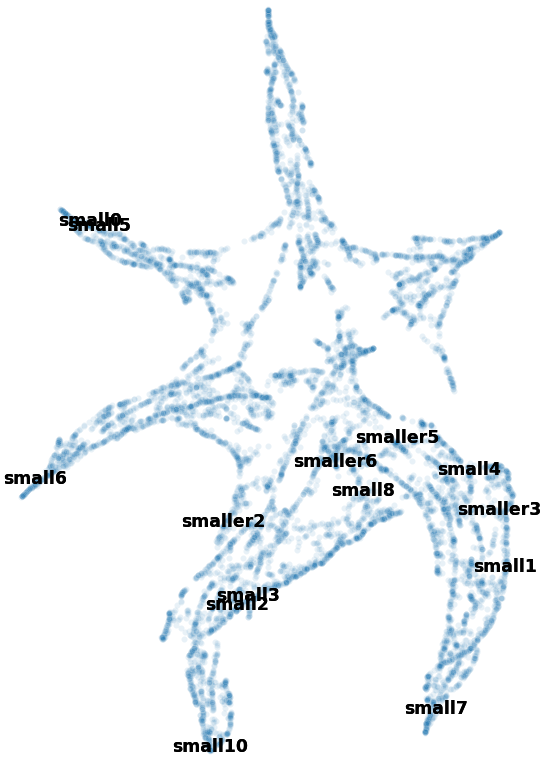}%
} 
\caption{Visualization of particular token locations on a space with 8 centroids mapped to 2D space with U-Map.}
\label{fig:qualitative2}
\end{figure*}

\begin{figure}
\centering

\subfigure[Space Division 1]{%
  \includegraphics[width=0.15\textwidth]{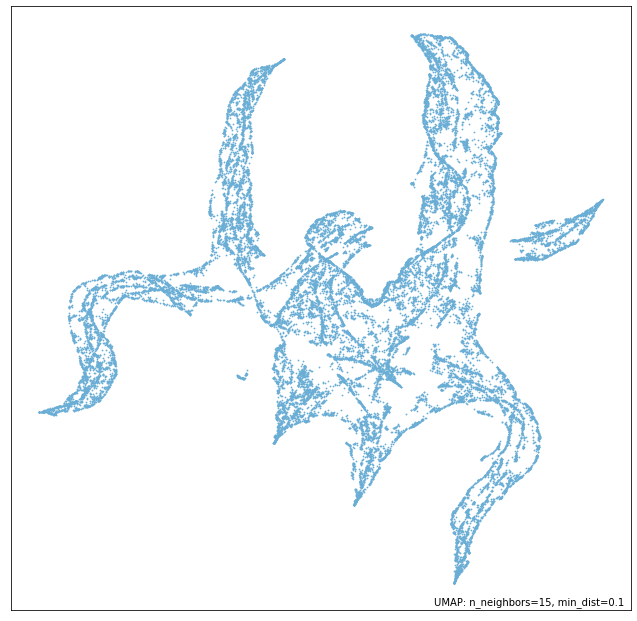}%
}
\subfigure[Space Division 2]{%
  \includegraphics[width=0.15\textwidth]{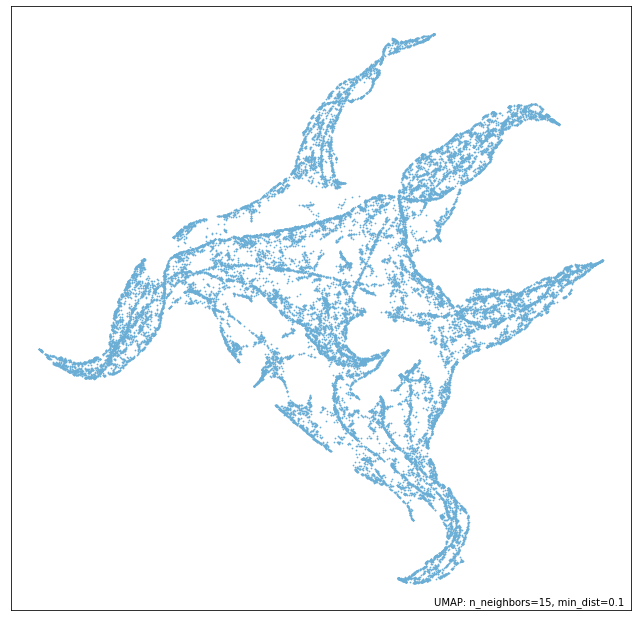}%
} 
\subfigure[Space Division 3]{%
  \includegraphics[width=0.15\textwidth]{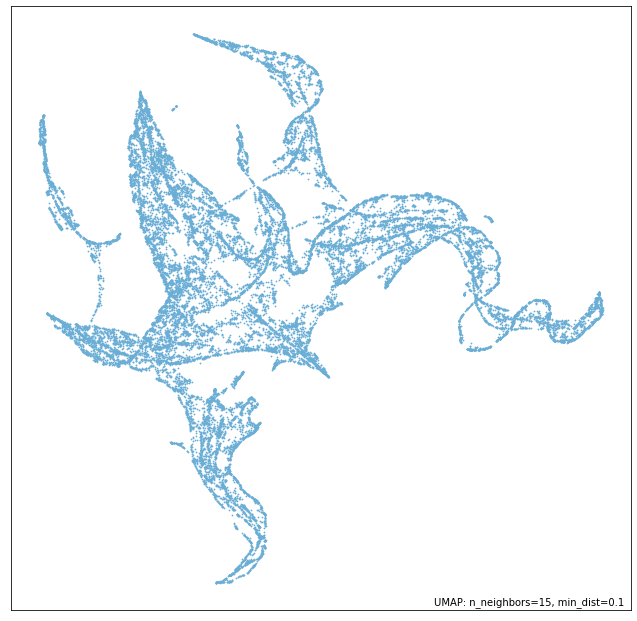}%
} 
\caption{Visualization of token sketches coming from 3 independent space divisions from a single T-EMDE coder, mapped to 2D with U-Map. Each space division features 8 centroids which are discernible in the pictures.}
\label{fig:qualitative1}
\end{figure}

\subsection{Results}

We present results on Flickr30k and MSCOCO 1K in Table \ref{tab:1k-results} and MSCOCO 5K in Table \ref{tab:5k-results}.  

We have retrained the original SAF and SGR models, as the original repository uses relatively old library versions -  Python v. 2.7, Pytorch v. 0.4.1, and NumPy v. 1.12.1. We have observed mostly higher results after bumping library versions to Python v. 3.6, Pytorch v. 1.7.1, NumPy v. 1.19.5. The original SAF/SGR results from \cite{Diao2021_sgraf} are denoted as \texttt{SAF/SGR reported} while our updated results are \texttt{SAF/SGR reproduced}. Additionally, we also report configurations without the Global Similarity Module as a baseline - \texttt{SAF/SGR no global}. All other competitors are recent top-performing models described in the Related Work Section. We only consider competitors which do not use external data, such as image segment location or segment caption, so that all models use exactly the same sources of information.

We can observe that T-EMDE brings significant improvements in most metrics, especially the R@1 metric (up to 1.7 pp.), in many cases establishing new state-of-the-art results, which leads us to think that it facilitates text/image matching while providing very fine-grained representations. The performance gains can be especially important in practical scenarios, as the R@1 metric denotes the top retrieved items which appear at the start of user recommendation list and determine the first user experience. 

The cases where T-EMDE gives non-SOTA results are usually at larger \textit{k} in Recall@k (which consider items from lower recommended ranks), and the performance difference to the top-performing model is usually small, from 0.1 to 0.3 pp. In particular, the non-SOTA results appear in T-EMDE/SGRAF MSCOCO 1K Image Retrieval. However, one needs to remember that the MSCOCO 1K evaluation protocol is the approximate version as it averages data chunks and ranking lists are computed on a smaller pool of candidates (a much easier task, hence the marked difference of scores between MSCOCO 1K and MSCOCO 5K which is seen in all models). On the full evaluation protocol - MSCOCO 5K - the T-EMDE models show considerable advantages in scores.
\section{Analysis}

\begin{figure}
\centering

\subfigure[R@1 Text Retrieval]{%
  \includegraphics[width=0.15\textwidth]{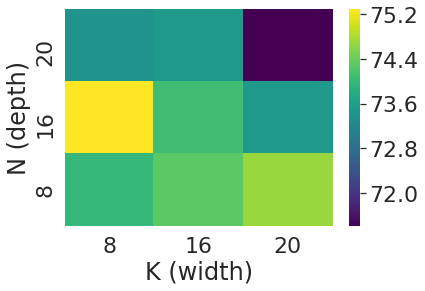}%
}
\subfigure[R@10 Text Retrieval]{%
  \includegraphics[width=0.15\textwidth]{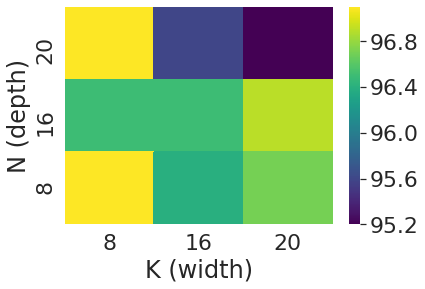}%
} 
\subfigure[MRR Text Retrieval]{%
  \includegraphics[width=0.15\textwidth]{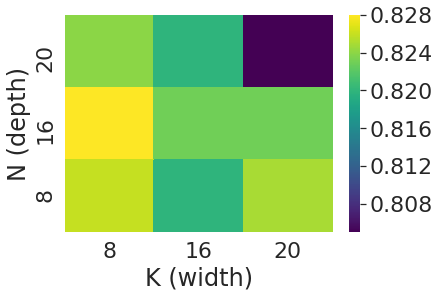}%
} 

% \\
\subfigure[R@1 Image Retrieval]{%
  \includegraphics[width=0.15\textwidth]{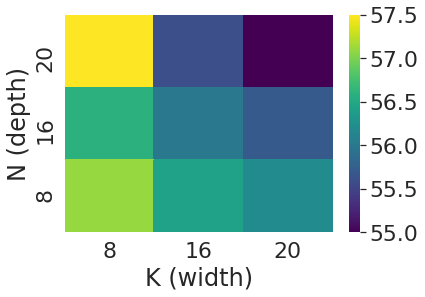}%
}
\subfigure[R@10 Image Retrieval]{%
  \includegraphics[width=0.15\textwidth]{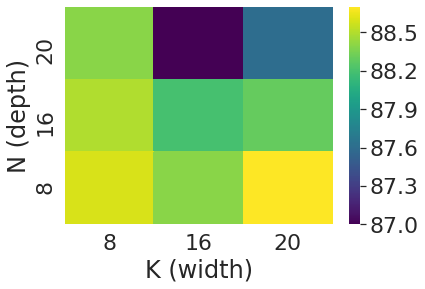}%
}
\subfigure[MRR Image Retrieval]{%
  \includegraphics[width=0.15\textwidth]{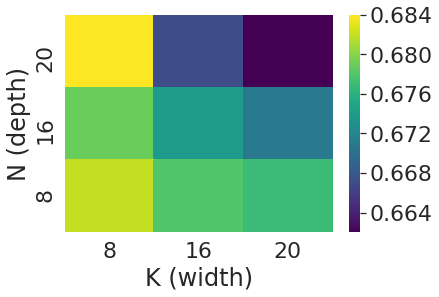}%
}
\caption{Performance scores on various N/K configurations on Flickr30K dataset.}
\label{fig:k-n-ablation}
\end{figure}

\begin{table}
\centering
\caption{Training and inference times of various configurations (averaged over 5 runs).}\label{tab:time}
\begin{tabular}{l|ll|ll|l} 
\hline
 & \multicolumn{2}{c}{\textbf{Flickr 30K}} &  \multicolumn{2}{c}{\textbf{COCO 1K}} & \multicolumn{1}{c}{\textbf{COCO 5K}}  \\
 Model & Infer & Train & Infer & Train & Infer \\\hline
 SAF & 69.68 s & 13h 35m & 68.1 s & 24h 54m & 1725.8 s \\
 T-EMDE & 59.68 s & 10h 23m & 58.1 s & 16h 45m & 1457.6 s \\
 \hline
 SGR &  83.6 s & 17h 35m & 84.0 s & 48h 52m & 2075.0 s \\
 T-EMDE & 75.4 s & 14h 01m & 74.9 s & 41h 22m & 1830.2 s \\\hline
 SGRAF & 161.4 s & - & 768.6 s & - & 4064.3 s\\
 T-EMDE & 135.5 s & - & 695.6 s & - & 3342.2 s\\
 \hline
\end{tabular}
\end{table}

\begin{table}
\centering
\caption{Performance of T-EMDE/SAF on Flickr30K with varying N/K configurations.}
\begin{tabular}{l|lll|lll|llll} 
\hline
 & \multicolumn{3}{c}{\textbf{R@1}} & \multicolumn{3}{c}{\textbf{R@10}} & \multicolumn{3}{c}{\textbf{MRR}}\\\hline
 N/K & 8 & 16 & 20 &  8 & 16 & 20 &  8 & 16 & 20  \\ \hline
 20 & 73.4 & 73.5 & 71.4 & 97.1 & 95.6 & 95.2 & 0.824 & 0.820 & 0.805 \\
 16 & 75.3 & 74.1 & 73.5 & 97.1 & 96.5 & 96.9 & 0.829 & 0.823 & 0.823\\
 8 & 74.0 & 74.3 & 74.7 & 97.1 & 96.4 & 96.7 & 0.826 & 0.820 & 0.825 \\
 \hline
\end{tabular}
\label{tab:ablation-res}
\end{table}

\subsection{Training and Inference Time}

As shown in Table \ref{tab:time}, T-EMDE makes both SAF and SGR faster by 15\%-20\% in every tested scenario in terms of inference time. This difference stems from reducing the quadratic complexity of the global similarity computed with self-attention to linear complexity of the trainable coder. Such performance difference can be significant in database retrieval scenarios, where low latency is of key importance. Likewise, training times are significantly reduced by a similar proportion as in the case of inference times. Note that the training speed difference does not stem from early stopping, as all benchmarked models were required to complete a defined number of epochs (20 epochs for MSCOCO, 30 epochs for Flickr30K SAF, 40 epochs for Flickr30K SGR).

Benchmarking was performed on a machine with 128 GB RAM, 14 core (28 HT threads) Intel Core i9-9940X 3.30GHz CPUs and GeForce RTX 2080 Ti 11GB GPU.

\subsection{Optimal N/K Configuration}
In Figure \ref{fig:k-n-ablation} and Table \ref{tab:ablation-res} we present R@1, R@10, and MRR performance scores for various $N$/$K$ parameter configurations of the T-EMDE/SAF model on the Flickr30K dataset. It can be observed that the T-EMDE sketch size can be indeed small, on the scale of $N \in \{8,16,20\}$ and $K=8$. This is consistent with observations from \cite{emde}: too large $K$ results in the loss of the information prior (too many centroids are created and even related items are assigned to different centroids). On the other hand, $N$ can often be large as its function is to introduce an ensembling effect. Independent space partitionings can thus be thought of as individual ensemble models. 

Overall, the resulting output sketch dimensionality can be dropped to $N=8$ and $K=8$, which produces a particularly small vector of size $8*8=64$, which is beneficial for model latency. At the same time, the performance loss from a very small coder can be considered low especially in practical scenarios (see Table \ref{tab:ablation-res}), which allows for further compression of the T-EMDE models.

\subsection{Qualitative Analysis}

Figure \ref{fig:qualitative1} shows 2D U-Map projections of token sketches obtained from  T-EMDE, coming from individual space partitionings. The evaluated T-EMDE coder contained 8 centroids, whose locations are discernible in the pictures. Particularly visible are boundary lines composed of individual tokens, akin to Voronoi cell boundaries. This suggests that many tokens are purposefully placed at equal distances to multiple centroids.

In Figure \ref{fig:qualitative2} we display the results of a 2D projection of a single 8-centroid space partitioning, but with annotated locations of particular tokens. It can be seen that various token classes occupy particular regions in space. For example, the words denoting sizes - \textit{big} and \textit{small} tend to group rather at the right bottom "legs" of the visualization. An interesting case are the  function words (exemplified with \textit{there} and \textit{these}) which are common in captions but bring in little semantic value. Such words are clearly grouped in a single cluster, which seems purposeful so that they can be ignored by the rest of the network.

\section{Conclusions}
In this paper we have presented T-EMDE - a highly efficient module for computing global representations of text and image in multimodal retrieval scenarios. By replacing a self-attentive module in recent high-performing architectures, we are able to reach new state-of-the-art results on MSCOCO and Flickr30k datasets, with especially significant gains in the Recall@1 metric. At the same time, we reduce the model latency by up to 20\%. We propose T-EMDE as a step in the direction of replacing computationally expensive self-attention with more efficient approaches.

\bibliographystyle{ACM-Reference-Format}
\bibliography{bibliography}


\begin{thebibliography}{62}

%%% the Web address.

\ifx \showCODEN    \undefined \def \showCODEN     #1{\unskip}     \fi
\ifx \showDOI      \undefined \def \showDOI       #1{#1}\fi
\ifx \showISBNx    \undefined \def \showISBNx     #1{\unskip}     \fi
\ifx \showISBNxiii \undefined \def \showISBNxiii  #1{\unskip}     \fi
\ifx \showISSN     \undefined \def \showISSN      #1{\unskip}     \fi
\ifx \showLCCN     \undefined \def \showLCCN      #1{\unskip}     \fi
\ifx \shownote     \undefined \def \shownote      #1{#1}          \fi
\ifx \showarticletitle \undefined \def \showarticletitle #1{#1}   \fi
\ifx \showURL      \undefined \def \showURL       {\relax}        \fi
% invisible to TeX
\providecommand\bibfield[2]{#2}
\providecommand\bibinfo[2]{#2}
\providecommand\natexlab[1]{#1}
\providecommand\showeprint[2][]{arXiv:#2}

\bibitem[\protect\citeauthoryear{Anderson, He, Buehler, Teney, Johnson, Gould,
  and Zhang}{Anderson et~al\mbox{.}}{2018}]%
        {DBLP:conf/cvpr/00010BT0GZ18}
\bibfield{author}{\bibinfo{person}{Peter Anderson}, \bibinfo{person}{Xiaodong
  He}, \bibinfo{person}{Chris Buehler}, \bibinfo{person}{Damien Teney},
  \bibinfo{person}{Mark Johnson}, \bibinfo{person}{Stephen Gould}, {and}
  \bibinfo{person}{Lei Zhang}.} \bibinfo{year}{2018}\natexlab{}.
\newblock \showarticletitle{Bottom-Up and Top-Down Attention for Image
  Captioning and Visual Question Answering}. In \bibinfo{booktitle}{\emph{2018
  {IEEE} Conference on Computer Vision and Pattern Recognition, {CVPR} 2018,
  Salt Lake City, UT, USA, June 18-22, 2018}}. \bibinfo{publisher}{{IEEE}
  Computer Society}, \bibinfo{pages}{6077--6086}.
\newblock
\urldef\tempurl%
\url{https://doi.org/10.1109/CVPR.2018.00636}
\showDOI{\tempurl}


\bibitem[\protect\citeauthoryear{Bahdanau, Cho, and Bengio}{Bahdanau
  et~al\mbox{.}}{2015}]%
        {Bahdanau2015_Neural_machine_translation_by_jointly}
\bibfield{author}{\bibinfo{person}{Dzmitry Bahdanau},
  \bibinfo{person}{Kyung~Hyun Cho}, {and} \bibinfo{person}{Yoshua Bengio}.}
  \bibinfo{year}{2015}\natexlab{}.
\newblock \showarticletitle{{Neural machine translation by jointly learning to
  align and translate}}. In \bibinfo{booktitle}{\emph{3rd International
  Conference on Learning Representations, ICLR 2015 - Conference Track
  Proceedings}}. \bibinfo{publisher}{International Conference on Learning
  Representations, ICLR}.
\newblock
\showeprint[arxiv]{1409.0473}
\urldef\tempurl%
\url{https://arxiv.org/abs/1409.0473v7}
\showURL{%
\tempurl}


\bibitem[\protect\citeauthoryear{Bai, Ren, and Zhang}{Bai
  et~al\mbox{.}}{2020}]%
        {Bai2020RippleWT}
\bibfield{author}{\bibinfo{person}{Jiyang Bai}, \bibinfo{person}{Yuxiang Ren},
  {and} \bibinfo{person}{J. Zhang}.} \bibinfo{year}{2020}\natexlab{}.
\newblock \showarticletitle{Ripple Walk Training: A Subgraph-based training
  framework for Large and Deep Graph Neural Network}.
\newblock \bibinfo{journal}{\emph{ArXiv}}  \bibinfo{volume}{abs/2002.07206}
  (\bibinfo{year}{2020}).
\newblock


\bibitem[\protect\citeauthoryear{Bello}{Bello}{[n.d.]}]%
        {bello2021lambdanetworks}
\bibfield{author}{\bibinfo{person}{Irwan Bello}.}
  \bibinfo{year}{[n.d.]}\natexlab{}.
\newblock \showarticletitle{LambdaNetworks: Modeling long-range Interactions
  without Attention}. In \bibinfo{booktitle}{\emph{International Conference on
  Learning Representations (ICLR) 2021}}.
\newblock


\bibitem[\protect\citeauthoryear{Beltagy, Peters, and Cohan}{Beltagy
  et~al\mbox{.}}{2020}]%
        {Beltagy_LongFormer}
\bibfield{author}{\bibinfo{person}{Iz Beltagy}, \bibinfo{person}{Matthew~E.
  Peters}, {and} \bibinfo{person}{Arman Cohan}.}
  \bibinfo{year}{2020}\natexlab{}.
\newblock \bibinfo{booktitle}{\emph{{Longformer: The Long-Document
  Transformer}}}.
\newblock \bibinfo{type}{{T}echnical {R}eport}.
\newblock
\showISSN{23318422}
\showeprint[arxiv]{2004.05150}
\urldef\tempurl%
\url{https://github.com/allenai/longformer}
\showURL{%
\tempurl}


\bibitem[\protect\citeauthoryear{Chen, Ding, Liu, Lin, Liu, and Han}{Chen
  et~al\mbox{.}}{2020}]%
        {Chen2020_IMRAM}
\bibfield{author}{\bibinfo{person}{Hui Chen}, \bibinfo{person}{Guiguang Ding},
  \bibinfo{person}{Xudong Liu}, \bibinfo{person}{Zijia Lin},
  \bibinfo{person}{Ji Liu}, {and} \bibinfo{person}{Jungong Han}.}
  \bibinfo{year}{2020}\natexlab{}.
\newblock \showarticletitle{{IMRAM: Iterative Matching with Recurrent Attention
  Memory for Cross-Modal Image-Text Retrieval}}.
\newblock \bibinfo{journal}{\emph{Proceedings of the IEEE Computer Society
  Conference on Computer Vision and Pattern Recognition}} (\bibinfo{date}{mar}
  \bibinfo{year}{2020}), \bibinfo{pages}{12652--12660}.
\newblock
\showISSN{10636919}
\urldef\tempurl%
\url{https://doi.org/10.1109/CVPR42600.2020.01267}
\showDOI{\tempurl}
\showeprint[arxiv]{2003.03772}


\bibitem[\protect\citeauthoryear{Chen, Zhu, and Song}{Chen
  et~al\mbox{.}}{[n.d.]}]%
        {pmlr-v80-chen18p}
\bibfield{author}{\bibinfo{person}{Jianfei Chen}, \bibinfo{person}{Jun Zhu},
  {and} \bibinfo{person}{Le Song}.} \bibinfo{year}{[n.d.]}\natexlab{}.
\newblock \showarticletitle{Stochastic Training of Graph Convolutional Networks
  with Variance Reduction}. In \bibinfo{booktitle}{\emph{Proceedings of the
  35th International Conference on Machine Learning}}.
\newblock


\bibitem[\protect\citeauthoryear{Chen and Luo}{Chen and Luo}{2020}]%
        {Chen2020a_Expressing_Objects_just_like_Words}
\bibfield{author}{\bibinfo{person}{Tianlang Chen} {and} \bibinfo{person}{Jiebo
  Luo}.} \bibinfo{year}{2020}\natexlab{}.
\newblock \showarticletitle{{Expressing objects just like words: Recurrent
  visual embedding for image-text matching}}.
\newblock \bibinfo{journal}{\emph{arXiv}} (\bibinfo{date}{feb}
  \bibinfo{year}{2020}).
\newblock
\showISSN{23318422}
\urldef\tempurl%
\url{https://doi.org/10.1609/aaai.v34i07.6631}
\showDOI{\tempurl}
\showeprint[arxiv]{2002.08510}


\bibitem[\protect\citeauthoryear{Child, Gray, Radford, and Sutskever}{Child
  et~al\mbox{.}}{2019}]%
        {Child_sparseTransfomer}
\bibfield{author}{\bibinfo{person}{Rewon Child}, \bibinfo{person}{Scott Gray},
  \bibinfo{person}{Alec Radford}, {and} \bibinfo{person}{Ilya Sutskever}.}
  \bibinfo{year}{2019}\natexlab{}.
\newblock \bibinfo{booktitle}{\emph{{Generating long sequences with sparse
  transformers}}}.
\newblock \bibinfo{type}{{T}echnical {R}eport}.
\newblock
\showISSN{23318422}
\showeprint[arxiv]{1904.10509}
\urldef\tempurl%
\url{https://openai.com/blog/sparse-transformer}
\showURL{%
\tempurl}


\bibitem[\protect\citeauthoryear{Cho, van Merri{\"e}nboer, Bahdanau, and
  Bengio}{Cho et~al\mbox{.}}{2014}]%
        {cho-etal-2014-properties}
\bibfield{author}{\bibinfo{person}{Kyunghyun Cho}, \bibinfo{person}{Bart van
  Merri{\"e}nboer}, \bibinfo{person}{Dzmitry Bahdanau}, {and}
  \bibinfo{person}{Yoshua Bengio}.} \bibinfo{year}{2014}\natexlab{}.
\newblock \showarticletitle{On the Properties of Neural Machine Translation:
  Encoder{--}Decoder Approaches}. In \bibinfo{booktitle}{\emph{Proceedings of
  {SSST}-8, Eighth Workshop on Syntax, Semantics and Structure in Statistical
  Translation}}. \bibinfo{publisher}{Association for Computational
  Linguistics}, \bibinfo{address}{Doha, Qatar}, \bibinfo{pages}{103--111}.
\newblock
\urldef\tempurl%
\url{https://doi.org/10.3115/v1/W14-4012}
\showDOI{\tempurl}


\bibitem[\protect\citeauthoryear{Choromanski, Likhosherstov, Dohan, Song, Gane,
  Sarlos, Hawkins, Davis, Mohiuddin, Kaiser, Belanger, Colwell, and
  Weller}{Choromanski et~al\mbox{.}}{2020}]%
        {Choromanski_PerFormer}
\bibfield{author}{\bibinfo{person}{Krzysztof Choromanski},
  \bibinfo{person}{Valerii Likhosherstov}, \bibinfo{person}{David Dohan},
  \bibinfo{person}{Xingyou Song}, \bibinfo{person}{Andreea Gane},
  \bibinfo{person}{Tamas Sarlos}, \bibinfo{person}{Peter Hawkins},
  \bibinfo{person}{Jared Davis}, \bibinfo{person}{Afroz Mohiuddin},
  \bibinfo{person}{Lukasz Kaiser}, \bibinfo{person}{David Belanger},
  \bibinfo{person}{Lucy Colwell}, {and} \bibinfo{person}{Adrian Weller}.}
  \bibinfo{year}{2020}\natexlab{}.
\newblock \bibinfo{booktitle}{\emph{{Rethinking attention with performers}}}.
\newblock \bibinfo{type}{{T}echnical {R}eport}.
\newblock
\showISSN{23318422}
\showeprint[arxiv]{2009.14794}


\bibitem[\protect\citeauthoryear{Chung, Gulcehre, Cho, and Bengio}{Chung
  et~al\mbox{.}}{2014}]%
        {Chung2014_empiricalEvaluationGRU}
\bibfield{author}{\bibinfo{person}{Junyoung Chung}, \bibinfo{person}{Caglar
  Gulcehre}, \bibinfo{person}{KyungHyun Cho}, {and} \bibinfo{person}{Yoshua
  Bengio}.} \bibinfo{year}{2014}\natexlab{}.
\newblock \bibinfo{booktitle}{\emph{{Empirical Evaluation of Gated Recurrent
  Neural Networks on Sequence Modeling}}}.
\newblock \bibinfo{type}{{T}echnical {R}eport}.
\newblock
\showeprint[arxiv]{1412.3555}
\urldef\tempurl%
\url{http://arxiv.org/abs/1412.3555}
\showURL{%
\tempurl}


\bibitem[\protect\citeauthoryear{Dai, Yang, Yang, Carbonell, Le, and
  Salakhutdinov}{Dai et~al\mbox{.}}{2020}]%
        {Dai_transformerXL}
\bibfield{author}{\bibinfo{person}{Zihang Dai}, \bibinfo{person}{Zhilin Yang},
  \bibinfo{person}{Yiming Yang}, \bibinfo{person}{Jaime Carbonell},
  \bibinfo{person}{Quoc~V. Le}, {and} \bibinfo{person}{Ruslan Salakhutdinov}.}
  \bibinfo{year}{2020}\natexlab{}.
\newblock \bibinfo{booktitle}{\emph{{Transformer-XL: Attentive language models
  beyond a fixed-length context}}}.
\newblock \bibinfo{type}{{T}echnical {R}eport}. \bibinfo{pages}{2978--2988}
  pages.
\newblock
\showISBNx{9781950737482}
\urldef\tempurl%
\url{https://doi.org/10.18653/v1/p19-1285}
\showDOI{\tempurl}
\showeprint[arxiv]{1901.02860}


\bibitem[\protect\citeauthoryear{Deshpande, Rasin, Brown, Furst, Raicu,
  Montner, and III}{Deshpande et~al\mbox{.}}{2017}]%
        {pmlr-v69-deshpande17a}
\bibfield{author}{\bibinfo{person}{Priya Deshpande}, \bibinfo{person}{Alexander
  Rasin}, \bibinfo{person}{Eli Brown}, \bibinfo{person}{Jacob Furst},
  \bibinfo{person}{Daniela Raicu}, \bibinfo{person}{Steven Montner}, {and}
  \bibinfo{person}{Samuel~Armato III}.} \bibinfo{year}{2017}\natexlab{}.
\newblock \showarticletitle{An Integrated Database and Smart Search Tool for
  Medical Knowledge Extraction from Radiology Teaching Files}. In
  \bibinfo{booktitle}{\emph{Proceedings of The First Workshop Medical
  Informatics and Healthcare held with the 23rd SIGKDD Conference on Knowledge
  Discovery and Data Mining}} \emph{(\bibinfo{series}{Proceedings of Machine
  Learning Research}, Vol.~\bibinfo{volume}{69})},
  \bibfield{editor}{\bibinfo{person}{Samah Fodeh} {and}
  \bibinfo{person}{Daniela~Stan Raicu}} (Eds.). \bibinfo{publisher}{PMLR},
  \bibinfo{pages}{10--18}.
\newblock
\urldef\tempurl%
\url{http://proceedings.mlr.press/v69/deshpande17a.html}
\showURL{%
\tempurl}


\bibitem[\protect\citeauthoryear{Diao, Zhang, Ma, and Lu}{Diao
  et~al\mbox{.}}{[n.d.]}]%
        {Diao2021_sgraf}
\bibfield{author}{\bibinfo{person}{Haiwen Diao}, \bibinfo{person}{Ying Zhang},
  \bibinfo{person}{Lin Ma}, {and} \bibinfo{person}{Huchuan Lu}.}
  \bibinfo{year}{[n.d.]}\natexlab{}.
\newblock \showarticletitle{{Similarity Reasoning and Filtration for Image-Text
  Matching}}. In \bibinfo{booktitle}{\emph{AAAI 2021}}.
\newblock


\bibitem[\protect\citeauthoryear{Dosovitskiy, Beyer, Kolesnikov, Weissenborn,
  Zhai, Unterthiner, Dehghani, Minderer, Heigold, Gelly, Uszkoreit, and
  Houlsby}{Dosovitskiy et~al\mbox{.}}{2020}]%
        {Dosovitskiy2020_image16words}
\bibfield{author}{\bibinfo{person}{Alexey Dosovitskiy}, \bibinfo{person}{Lucas
  Beyer}, \bibinfo{person}{Alexander Kolesnikov}, \bibinfo{person}{Dirk
  Weissenborn}, \bibinfo{person}{Xiaohua Zhai}, \bibinfo{person}{Thomas
  Unterthiner}, \bibinfo{person}{Mostafa Dehghani}, \bibinfo{person}{Matthias
  Minderer}, \bibinfo{person}{Georg Heigold}, \bibinfo{person}{Sylvain Gelly},
  \bibinfo{person}{Jakob Uszkoreit}, {and} \bibinfo{person}{Neil Houlsby}.}
  \bibinfo{year}{2020}\natexlab{}.
\newblock \showarticletitle{{An Image is Worth 16x16 Words: Transformers for
  Image Recognition at Scale}}.
\newblock  (\bibinfo{date}{oct} \bibinfo{year}{2020}).
\newblock
\showeprint[arxiv]{2010.11929}
\urldef\tempurl%
\url{http://arxiv.org/abs/2010.11929}
\showURL{%
\tempurl}


\bibitem[\protect\citeauthoryear{Dąbrowski, Rychalska, Daniluk, Basaj, Babel,
  Michałowski, and Jakubowski}{Dąbrowski et~al\mbox{.}}{2021}]%
        {emde}
\bibfield{author}{\bibinfo{person}{Jacek Dąbrowski}, \bibinfo{person}{Barbara
  Rychalska}, \bibinfo{person}{Michał Daniluk}, \bibinfo{person}{Dominika
  Basaj}, \bibinfo{person}{Piotr Babel}, \bibinfo{person}{Andrzej
  Michałowski}, {and} \bibinfo{person}{Adam Jakubowski}.}
  \bibinfo{year}{2021}\natexlab{}.
\newblock \showarticletitle{An efficient manifold density estimator for all
  recommendation systems}.
\newblock
\urldef\tempurl%
\url{https://arxiv.org/abs/2006.01894}
\showURL{%
\tempurl}


\bibitem[\protect\citeauthoryear{Elizalde, Zarar, and Raj}{Elizalde
  et~al\mbox{.}}{2019}]%
        {8682632}
\bibfield{author}{\bibinfo{person}{Benjamin Elizalde}, \bibinfo{person}{Shuayb
  Zarar}, {and} \bibinfo{person}{Bhiksha Raj}.}
  \bibinfo{year}{2019}\natexlab{}.
\newblock \showarticletitle{Cross Modal Audio Search and Retrieval with Joint
  Embeddings Based on Text and Audio}. In \bibinfo{booktitle}{\emph{ICASSP 2019
  - 2019 IEEE International Conference on Acoustics, Speech and Signal
  Processing (ICASSP)}}. \bibinfo{pages}{4095--4099}.
\newblock
\urldef\tempurl%
\url{https://doi.org/10.1109/ICASSP.2019.8682632}
\showDOI{\tempurl}


\bibitem[\protect\citeauthoryear{Faghri, Fleet, Kiros, and Fidler}{Faghri
  et~al\mbox{.}}{2017}]%
        {Faghri2017_VSEpp}
\bibfield{author}{\bibinfo{person}{Fartash Faghri}, \bibinfo{person}{David~J.
  Fleet}, \bibinfo{person}{Jamie~Ryan Kiros}, {and} \bibinfo{person}{Sanja
  Fidler}.} \bibinfo{year}{2017}\natexlab{}.
\newblock \showarticletitle{{VSE++: Improving visual-semantic embeddings with
  hard negatives}}.
\newblock \bibinfo{journal}{\emph{arXiv}} (\bibinfo{date}{jul}
  \bibinfo{year}{2017}).
\newblock
\showISSN{23318422}
\showeprint[arxiv]{1707.05612}
\urldef\tempurl%
\url{http://arxiv.org/abs/1707.05612}
\showURL{%
\tempurl}


\bibitem[\protect\citeauthoryear{Frome, Corrado, Shlens, Bengio, Dean, Ranzato,
  and Mikolov}{Frome et~al\mbox{.}}{2013}]%
        {NIPS2013_7cce53cf}
\bibfield{author}{\bibinfo{person}{Andrea Frome}, \bibinfo{person}{Greg~S
  Corrado}, \bibinfo{person}{Jon Shlens}, \bibinfo{person}{Samy Bengio},
  \bibinfo{person}{Jeff Dean}, \bibinfo{person}{Marc\textquotesingle~Aurelio
  Ranzato}, {and} \bibinfo{person}{Tomas Mikolov}.}
  \bibinfo{year}{2013}\natexlab{}.
\newblock \showarticletitle{DeViSE: A Deep Visual-Semantic Embedding Model}. In
  \bibinfo{booktitle}{\emph{Advances in Neural Information Processing
  Systems}}, \bibfield{editor}{\bibinfo{person}{C.~J.~C. Burges},
  \bibinfo{person}{L.~Bottou}, \bibinfo{person}{M.~Welling},
  \bibinfo{person}{Z.~Ghahramani}, {and} \bibinfo{person}{K.~Q. Weinberger}}
  (Eds.), Vol.~\bibinfo{volume}{26}. \bibinfo{publisher}{Curran Associates,
  Inc.}
\newblock
\urldef\tempurl%
\url{https://proceedings.neurips.cc/paper/2013/file/7cce53cf90577442771720a370c3c723-Paper.pdf}
\showURL{%
\tempurl}


\bibitem[\protect\citeauthoryear{Guan, Liu, Yan, Qian, and Ji}{Guan
  et~al\mbox{.}}{2018}]%
        {coco_flickr_comparison}
\bibfield{author}{\bibinfo{person}{Zhibin Guan}, \bibinfo{person}{Kang Liu},
  \bibinfo{person}{Ma Yan}, \bibinfo{person}{Xu Qian}, {and}
  \bibinfo{person}{Tongkai Ji}.} \bibinfo{year}{2018}\natexlab{}.
\newblock \showarticletitle{Sequential Dual Attention: Coarse-to-Fine-Grained
  Hierarchical Generation for Image Captioning}.
\newblock \bibinfo{journal}{\emph{Symmetry}}  \bibinfo{volume}{10}
  (\bibinfo{date}{11} \bibinfo{year}{2018}), \bibinfo{pages}{626}.
\newblock
\urldef\tempurl%
\url{https://doi.org/10.3390/sym10110626}
\showDOI{\tempurl}


\bibitem[\protect\citeauthoryear{Hossain, Sohel, Shiratuddin, and Laga}{Hossain
  et~al\mbox{.}}{2019}]%
        {10.1145/3295748}
\bibfield{author}{\bibinfo{person}{MD.~Zakir Hossain}, \bibinfo{person}{Ferdous
  Sohel}, \bibinfo{person}{Mohd~Fairuz Shiratuddin}, {and}
  \bibinfo{person}{Hamid Laga}.} \bibinfo{year}{2019}\natexlab{}.
\newblock \showarticletitle{A Comprehensive Survey of Deep Learning for Image
  Captioning}.
\newblock \bibinfo{journal}{\emph{ACM Comput. Surv.}} \bibinfo{volume}{51},
  \bibinfo{number}{6}, Article \bibinfo{articleno}{118} (\bibinfo{date}{Feb.}
  \bibinfo{year}{2019}), \bibinfo{numpages}{36}~pages.
\newblock
\showISSN{0360-0300}
\urldef\tempurl%
\url{https://doi.org/10.1145/3295748}
\showDOI{\tempurl}


\bibitem[\protect\citeauthoryear{Hu, Luo, Lin, Yan, and Chen}{Hu
  et~al\mbox{.}}{2019}]%
        {Hu2019_Multi-level_visual-semantic}
\bibfield{author}{\bibinfo{person}{Zhibin Hu}, \bibinfo{person}{Yongsheng Luo},
  \bibinfo{person}{Jiong Lin}, \bibinfo{person}{Yan~Y. Yan}, {and}
  \bibinfo{person}{Jian Chen}.} \bibinfo{year}{2019}\natexlab{}.
\newblock \showarticletitle{{Multi-level visual-semantic alignments with
  relation-wise dual attention network for image and text matching}}. In
  \bibinfo{booktitle}{\emph{IJCAI International Joint Conference on Artificial
  Intelligence}}, Vol.~\bibinfo{volume}{2019-August}.
  \bibinfo{publisher}{International Joint Conferences on Artificial
  Intelligence}, \bibinfo{pages}{789--795}.
\newblock
\showISBNx{9780999241141}
\showISSN{10450823}
\urldef\tempurl%
\url{https://doi.org/10.24963/ijcai.2019/111}
\showDOI{\tempurl}


\bibitem[\protect\citeauthoryear{Huang, Wang, Chen, and Wei}{Huang
  et~al\mbox{.}}{2019}]%
        {Huang_2019_ICCV}
\bibfield{author}{\bibinfo{person}{Lun Huang}, \bibinfo{person}{Wenmin Wang},
  \bibinfo{person}{Jie Chen}, {and} \bibinfo{person}{Xiao-Yong Wei}.}
  \bibinfo{year}{2019}\natexlab{}.
\newblock \showarticletitle{Attention on Attention for Image Captioning}. In
  \bibinfo{booktitle}{\emph{Proceedings of the IEEE/CVF International
  Conference on Computer Vision (ICCV)}}.
\newblock


\bibitem[\protect\citeauthoryear{Huang, Chang, Hauptmann, and Hovy}{Huang
  et~al\mbox{.}}{2020}]%
        {10.1145/3372278.3390674}
\bibfield{author}{\bibinfo{person}{Po-Yao Huang}, \bibinfo{person}{Xiaojun
  Chang}, \bibinfo{person}{Alexander Hauptmann}, {and} \bibinfo{person}{Eduard
  Hovy}.} \bibinfo{year}{2020}\natexlab{}.
\newblock \showarticletitle{Forward and Backward Multimodal NMT for Improved
  Monolingual and Multilingual Cross-Modal Retrieval}. In
  \bibinfo{booktitle}{\emph{Proceedings of the 2020 International Conference on
  Multimedia Retrieval}} (Dublin, Ireland) \emph{(\bibinfo{series}{ICMR '20})}.
  \bibinfo{publisher}{Association for Computing Machinery},
  \bibinfo{address}{New York, NY, USA}, \bibinfo{pages}{53–62}.
\newblock
\showISBNx{9781450370875}
\urldef\tempurl%
\url{https://doi.org/10.1145/3372278.3390674}
\showDOI{\tempurl}


\bibitem[\protect\citeauthoryear{Karpathy and Fei-Fei}{Karpathy and
  Fei-Fei}{2017}]%
        {Karpathy2014}
\bibfield{author}{\bibinfo{person}{Andrej Karpathy} {and} \bibinfo{person}{Li
  Fei-Fei}.} \bibinfo{year}{2017}\natexlab{}.
\newblock \showarticletitle{{Deep Visual-Semantic Alignments for Generating
  Image Descriptions}}.
\newblock \bibinfo{journal}{\emph{IEEE Transactions on Pattern Analysis and
  Machine Intelligence}} \bibinfo{volume}{39}, \bibinfo{number}{4}
  (\bibinfo{date}{dec} \bibinfo{year}{2017}), \bibinfo{pages}{664--676}.
\newblock
\showISSN{01628828}
\urldef\tempurl%
\url{https://doi.org/10.1109/TPAMI.2016.2598339}
\showDOI{\tempurl}
\showeprint[arxiv]{1412.2306}


\bibitem[\protect\citeauthoryear{Katharopoulos, Vyas, Pappas, and
  Fleuret}{Katharopoulos et~al\mbox{.}}{2020}]%
        {Katharopoulos2020_TransformerAreRNNs}
\bibfield{author}{\bibinfo{person}{Angelos Katharopoulos},
  \bibinfo{person}{Apoorv Vyas}, \bibinfo{person}{Nikolaos Pappas}, {and}
  \bibinfo{person}{Fran{\c{c}}ois Fleuret}.} \bibinfo{year}{2020}\natexlab{}.
\newblock \showarticletitle{{Transformers are RNNs: Fast Autoregressive
  Transformers with Linear Attention}}.
\newblock \bibinfo{journal}{\emph{arXiv}} (\bibinfo{date}{jun}
  \bibinfo{year}{2020}).
\newblock
\showISSN{23318422}
\showeprint[arxiv]{2006.16236}
\urldef\tempurl%
\url{http://arxiv.org/abs/2006.16236}
\showURL{%
\tempurl}


\bibitem[\protect\citeauthoryear{Kitaev, Kaiser, and Levskaya}{Kitaev
  et~al\mbox{.}}{2020}]%
        {Kitaev_ReFormer}
\bibfield{author}{\bibinfo{person}{Nikita Kitaev}, \bibinfo{person}{{\L}ukasz
  Kaiser}, {and} \bibinfo{person}{Anselm Levskaya}.}
  \bibinfo{year}{2020}\natexlab{}.
\newblock \bibinfo{booktitle}{\emph{{Reformer: the Efficient Transformer}}}.
\newblock \bibinfo{type}{{T}echnical {R}eport}.
\newblock
\showISSN{23318422}
\showeprint[arxiv]{2001.04451}
\urldef\tempurl%
\url{https://hackingsemantics.xyz/2019/leaderboards/}
\showURL{%
\tempurl}


\bibitem[\protect\citeauthoryear{Krishna, Zhu, Groth, Johnson, Hata, Kravitz,
  Chen, Kalantidis, Li, Shamma, Bernstein, and Fei{-}Fei}{Krishna
  et~al\mbox{.}}{2017}]%
        {DBLP:journals/ijcv/KrishnaZGJHKCKL17}
\bibfield{author}{\bibinfo{person}{Ranjay Krishna}, \bibinfo{person}{Yuke Zhu},
  \bibinfo{person}{Oliver Groth}, \bibinfo{person}{Justin Johnson},
  \bibinfo{person}{Kenji Hata}, \bibinfo{person}{Joshua Kravitz},
  \bibinfo{person}{Stephanie Chen}, \bibinfo{person}{Yannis Kalantidis},
  \bibinfo{person}{Li{-}Jia Li}, \bibinfo{person}{David~A. Shamma},
  \bibinfo{person}{Michael~S. Bernstein}, {and} \bibinfo{person}{Li
  Fei{-}Fei}.} \bibinfo{year}{2017}\natexlab{}.
\newblock \showarticletitle{Visual Genome: Connecting Language and Vision Using
  Crowdsourced Dense Image Annotations}.
\newblock \bibinfo{journal}{\emph{Int. J. Comput. Vis.}} \bibinfo{volume}{123},
  \bibinfo{number}{1} (\bibinfo{year}{2017}), \bibinfo{pages}{32--73}.
\newblock
\urldef\tempurl%
\url{https://doi.org/10.1007/s11263-016-0981-7}
\showDOI{\tempurl}


\bibitem[\protect\citeauthoryear{Kuang, Gao, Li, Luo, Chen, Lin, and
  Zhang}{Kuang et~al\mbox{.}}{2019}]%
        {DBLP:conf/iccv/Kuang0LLCLZ19}
\bibfield{author}{\bibinfo{person}{Zhanghui Kuang}, \bibinfo{person}{Yiming
  Gao}, \bibinfo{person}{Guanbin Li}, \bibinfo{person}{Ping Luo},
  \bibinfo{person}{Yimin Chen}, \bibinfo{person}{Liang Lin}, {and}
  \bibinfo{person}{Wayne Zhang}.} \bibinfo{year}{2019}\natexlab{}.
\newblock \showarticletitle{Fashion Retrieval via Graph Reasoning Networks on a
  Similarity Pyramid}. In \bibinfo{booktitle}{\emph{2019 {IEEE/CVF}
  International Conference on Computer Vision, {ICCV} 2019, Seoul, Korea
  (South), October 27 - November 2, 2019}}. \bibinfo{publisher}{{IEEE}},
  \bibinfo{pages}{3066--3075}.
\newblock
\urldef\tempurl%
\url{https://doi.org/10.1109/ICCV.2019.00316}
\showDOI{\tempurl}


\bibitem[\protect\citeauthoryear{Lee, Chen, Hua, Hu, and He}{Lee
  et~al\mbox{.}}{2018}]%
        {Lee2018_scan}
\bibfield{author}{\bibinfo{person}{Kuang~Huei Lee}, \bibinfo{person}{Xi Chen},
  \bibinfo{person}{Gang Hua}, \bibinfo{person}{Houdong Hu}, {and}
  \bibinfo{person}{Xiaodong He}.} \bibinfo{year}{2018}\natexlab{}.
\newblock \showarticletitle{{Stacked Cross Attention for Image-Text Matching}}.
\newblock \bibinfo{journal}{\emph{Lecture Notes in Computer Science (including
  subseries Lecture Notes in Artificial Intelligence and Lecture Notes in
  Bioinformatics)}}  \bibinfo{volume}{11208 LNCS} (\bibinfo{date}{mar}
  \bibinfo{year}{2018}), \bibinfo{pages}{212--228}.
\newblock
\showISBNx{9783030012243}
\showISSN{16113349}
\urldef\tempurl%
\url{https://doi.org/10.1007/978-3-030-01225-0_13}
\showDOI{\tempurl}
\showeprint[arxiv]{1803.08024}


\bibitem[\protect\citeauthoryear{Li, Zhang, Li, Li, and Fu}{Li
  et~al\mbox{.}}{2019}]%
        {Li2019_Visual_Semantic_Reasoning_for_Image-Text_Matching}
\bibfield{author}{\bibinfo{person}{Kunpeng Li}, \bibinfo{person}{Yulun Zhang},
  \bibinfo{person}{Kai Li}, \bibinfo{person}{Yuanyuan Li}, {and}
  \bibinfo{person}{Yun Fu}.} \bibinfo{year}{2019}\natexlab{}.
\newblock \showarticletitle{{Visual semantic reasoning for image-text
  matching}}.
\newblock \bibinfo{journal}{\emph{Proceedings of the IEEE International
  Conference on Computer Vision}}  \bibinfo{volume}{2019-October}
  (\bibinfo{date}{sep} \bibinfo{year}{2019}), \bibinfo{pages}{4653--4661}.
\newblock
\showISBNx{9781728148038}
\showISSN{15505499}
\urldef\tempurl%
\url{https://doi.org/10.1109/ICCV.2019.00475}
\showDOI{\tempurl}
\showeprint[arxiv]{1909.02701}


\bibitem[\protect\citeauthoryear{Li, Xiao, Li, Zhou, Yue, and Wang}{Li
  et~al\mbox{.}}{2017}]%
        {Li_2017_CVPR}
\bibfield{author}{\bibinfo{person}{Shuang Li}, \bibinfo{person}{Tong Xiao},
  \bibinfo{person}{Hongsheng Li}, \bibinfo{person}{Bolei Zhou},
  \bibinfo{person}{Dayu Yue}, {and} \bibinfo{person}{Xiaogang Wang}.}
  \bibinfo{year}{2017}\natexlab{}.
\newblock \showarticletitle{Person Search With Natural Language Description}.
  In \bibinfo{booktitle}{\emph{Proceedings of the IEEE Conference on Computer
  Vision and Pattern Recognition (CVPR)}}.
\newblock


\bibitem[\protect\citeauthoryear{Lin, Maire, Belongie, Hays, Perona, Ramanan,
  Doll{\'{a}}r, and Zitnick}{Lin et~al\mbox{.}}{2014}]%
        {DBLP:conf/eccv/LinMBHPRDZ14}
\bibfield{author}{\bibinfo{person}{Tsung{-}Yi Lin}, \bibinfo{person}{Michael
  Maire}, \bibinfo{person}{Serge~J. Belongie}, \bibinfo{person}{James Hays},
  \bibinfo{person}{Pietro Perona}, \bibinfo{person}{Deva Ramanan},
  \bibinfo{person}{Piotr Doll{\'{a}}r}, {and} \bibinfo{person}{C.~Lawrence
  Zitnick}.} \bibinfo{year}{2014}\natexlab{}.
\newblock \showarticletitle{Microsoft {COCO:} Common Objects in Context}. In
  \bibinfo{booktitle}{\emph{Computer Vision - {ECCV} 2014 - 13th European
  Conference, Zurich, Switzerland, September 6-12, 2014, Proceedings, Part
  {V}}} \emph{(\bibinfo{series}{Lecture Notes in Computer Science},
  Vol.~\bibinfo{volume}{8693})}, \bibfield{editor}{\bibinfo{person}{David~J.
  Fleet}, \bibinfo{person}{Tom{\'{a}}s Pajdla}, \bibinfo{person}{Bernt
  Schiele}, {and} \bibinfo{person}{Tinne Tuytelaars}} (Eds.).
  \bibinfo{publisher}{Springer}, \bibinfo{pages}{740--755}.
\newblock
\urldef\tempurl%
\url{https://doi.org/10.1007/978-3-319-10602-1\_48}
\showDOI{\tempurl}


\bibitem[\protect\citeauthoryear{Loper and Bird}{Loper and Bird}{2002}]%
        {Loper02nltk:the}
\bibfield{author}{\bibinfo{person}{Edward Loper} {and} \bibinfo{person}{Steven
  Bird}.} \bibinfo{year}{2002}\natexlab{}.
\newblock \showarticletitle{NLTK: The Natural Language Toolkit}. In
  \bibinfo{booktitle}{\emph{In Proceedings of the ACL Workshop on Effective
  Tools and Methodologies for Teaching Natural Language Processing and
  Computational Linguistics. Philadelphia: Association for Computational
  Linguistics}}.
\newblock


\bibitem[\protect\citeauthoryear{Nishihara, Tamura, Ninomiya, Omote, and
  Nakayama}{Nishihara et~al\mbox{.}}{2020}]%
        {nishihara-etal-2020-supervised}
\bibfield{author}{\bibinfo{person}{Tetsuro Nishihara}, \bibinfo{person}{Akihiro
  Tamura}, \bibinfo{person}{Takashi Ninomiya}, \bibinfo{person}{Yutaro Omote},
  {and} \bibinfo{person}{Hideki Nakayama}.} \bibinfo{year}{2020}\natexlab{}.
\newblock \showarticletitle{Supervised Visual Attention for Multimodal Neural
  Machine Translation}. In \bibinfo{booktitle}{\emph{Proceedings of the 28th
  International Conference on Computational Linguistics}}.
  \bibinfo{publisher}{International Committee on Computational Linguistics},
  \bibinfo{address}{Barcelona, Spain (Online)}, \bibinfo{pages}{4304--4314}.
\newblock
\urldef\tempurl%
\url{https://doi.org/10.18653/v1/2020.coling-main.380}
\showDOI{\tempurl}


\bibitem[\protect\citeauthoryear{Qiu, Ma, Levy, Yih, Wang, and Tang}{Qiu
  et~al\mbox{.}}{2019}]%
        {Qiu2019_BlockBERT}
\bibfield{author}{\bibinfo{person}{Jiezhong Qiu}, \bibinfo{person}{Hao Ma},
  \bibinfo{person}{Omer Levy}, \bibinfo{person}{Scott Wen~Tau Yih},
  \bibinfo{person}{Sinong Wang}, {and} \bibinfo{person}{Jie Tang}.}
  \bibinfo{year}{2019}\natexlab{}.
\newblock \showarticletitle{{Blockwise self-attention for long document
  understanding}}.
\newblock \bibinfo{journal}{\emph{arXiv}} (\bibinfo{date}{nov}
  \bibinfo{year}{2019}).
\newblock
\showISSN{23318422}
\urldef\tempurl%
\url{https://doi.org/10.18653/v1/2020.findings-emnlp.232}
\showDOI{\tempurl}
\showeprint[arxiv]{1911.02972}


\bibitem[\protect\citeauthoryear{Raffel, Shazeer, Roberts, Lee, Narang, Matena,
  Zhou, Peter, and Liu}{Raffel et~al\mbox{.}}{2019}]%
        {Raffel2020_Exploring_the_Limits_of_Transfer}
\bibfield{author}{\bibinfo{person}{Colin Raffel}, \bibinfo{person}{Noam
  Shazeer}, \bibinfo{person}{Adam Roberts}, \bibinfo{person}{Katherine Lee},
  \bibinfo{person}{Sharan Narang}, \bibinfo{person}{Michael Matena},
  \bibinfo{person}{Yanqi Zhou}, \bibinfo{person}{Wei~Li Peter}, {and}
  \bibinfo{person}{J. Liu}.} \bibinfo{year}{2019}\natexlab{}.
\newblock \bibinfo{booktitle}{\emph{{Exploring the limits of transfer learning
  with a unified text-to-text transformer}}}.
\newblock \bibinfo{type}{{T}echnical {R}eport}. \bibinfo{pages}{1--67} pages.
\newblock
\showISSN{23318422}
\showeprint[arxiv]{1910.10683}
\urldef\tempurl%
\url{http://jmlr.org/papers/v21/20-074.html.}
\showURL{%
\tempurl}


\bibitem[\protect\citeauthoryear{Ren, He, Girshick, and Sun}{Ren
  et~al\mbox{.}}{2015}]%
        {NIPS2015_14bfa6bb}
\bibfield{author}{\bibinfo{person}{Shaoqing Ren}, \bibinfo{person}{Kaiming He},
  \bibinfo{person}{Ross Girshick}, {and} \bibinfo{person}{Jian Sun}.}
  \bibinfo{year}{2015}\natexlab{}.
\newblock \showarticletitle{Faster R-CNN: Towards Real-Time Object Detection
  with Region Proposal Networks}. In \bibinfo{booktitle}{\emph{Advances in
  Neural Information Processing Systems}},
  \bibfield{editor}{\bibinfo{person}{C.~Cortes}, \bibinfo{person}{N.~Lawrence},
  \bibinfo{person}{D.~Lee}, \bibinfo{person}{M.~Sugiyama}, {and}
  \bibinfo{person}{R.~Garnett}} (Eds.), Vol.~\bibinfo{volume}{28}.
  \bibinfo{publisher}{Curran Associates, Inc.}
\newblock
\urldef\tempurl%
\url{https://proceedings.neurips.cc/paper/2015/file/14bfa6bb14875e45bba028a21ed38046-Paper.pdf}
\showURL{%
\tempurl}


\bibitem[\protect\citeauthoryear{Roy, Saffar, Vaswani, and Grangier}{Roy
  et~al\mbox{.}}{2020}]%
        {Roy2020_RoutingTransformer}
\bibfield{author}{\bibinfo{person}{Aurko Roy}, \bibinfo{person}{Mohammad
  Saffar}, \bibinfo{person}{Ashish Vaswani}, {and} \bibinfo{person}{David
  Grangier}.} \bibinfo{year}{2020}\natexlab{}.
\newblock \bibinfo{booktitle}{\emph{{Efficient content-based sparse attention
  with routing transformers}}}.
\newblock \bibinfo{type}{{T}echnical {R}eport}.
\newblock
\showISSN{23318422}
\urldef\tempurl%
\url{https://doi.org/10.1162/tacl_a_00353}
\showDOI{\tempurl}
\showeprint[arxiv]{2003.05997}


\bibitem[\protect\citeauthoryear{Song and Soleymani}{Song and
  Soleymani}{2019}]%
        {Song2019_Polysemous}
\bibfield{author}{\bibinfo{person}{Yale Song} {and} \bibinfo{person}{Mohammad
  Soleymani}.} \bibinfo{year}{2019}\natexlab{}.
\newblock \showarticletitle{{Polysemous visual-semantic embedding for
  cross-modal retrieval}}.
\newblock \bibinfo{journal}{\emph{Proceedings of the IEEE Computer Society
  Conference on Computer Vision and Pattern Recognition}}
  \bibinfo{volume}{2019-June} (\bibinfo{date}{jun} \bibinfo{year}{2019}),
  \bibinfo{pages}{1979--1988}.
\newblock
\showISBNx{9781728132938}
\showISSN{10636919}
\urldef\tempurl%
\url{https://doi.org/10.1109/CVPR.2019.00208}
\showDOI{\tempurl}
\showeprint[arxiv]{1906.04402}


\bibitem[\protect\citeauthoryear{Su, Zhu, Cao, Li, Lu, Wei, and Dai}{Su
  et~al\mbox{.}}{2019}]%
        {Su_VLBERT}
\bibfield{author}{\bibinfo{person}{Weijie Su}, \bibinfo{person}{Xizhou Zhu},
  \bibinfo{person}{Yue Cao}, \bibinfo{person}{Bin Li}, \bibinfo{person}{Lewei
  Lu}, \bibinfo{person}{Furu Wei}, {and} \bibinfo{person}{Jifeng Dai}.}
  \bibinfo{year}{2019}\natexlab{}.
\newblock \bibinfo{booktitle}{\emph{{VL-BERT: Pre-training of generic
  visual-linguistic representations}}}.
\newblock \bibinfo{type}{{T}echnical {R}eport}.
\newblock
\showISBNx{1908.08530v4}
\showISSN{23318422}
\showeprint[arxiv]{1908.08530}
\urldef\tempurl%
\url{https://github.com/jackroos/VL-BERT.}
\showURL{%
\tempurl}


\bibitem[\protect\citeauthoryear{Tan and Bansal}{Tan and Bansal}{2020}]%
        {Tan_LXMERT}
\bibfield{author}{\bibinfo{person}{Hao Tan} {and} \bibinfo{person}{Mohit
  Bansal}.} \bibinfo{year}{2020}\natexlab{}.
\newblock \bibinfo{booktitle}{\emph{{LXMert: Learning cross-modality encoder
  representations from transformers}}}.
\newblock \bibinfo{type}{{T}echnical {R}eport}. \bibinfo{pages}{5100--5111}
  pages.
\newblock
\showISBNx{9781950737901}
\urldef\tempurl%
\url{https://doi.org/10.18653/v1/d19-1514}
\showDOI{\tempurl}
\showeprint[arxiv]{1908.07490}


\bibitem[\protect\citeauthoryear{Tay, Bahri, Yang, Metzler, and Juan}{Tay
  et~al\mbox{.}}{2020a}]%
        {Tay2020_Sinkhorn}
\bibfield{author}{\bibinfo{person}{Yi Tay}, \bibinfo{person}{Dara Bahri},
  \bibinfo{person}{Liu Yang}, \bibinfo{person}{Donald Metzler}, {and}
  \bibinfo{person}{Da~Cheng Juan}.} \bibinfo{year}{2020}\natexlab{a}.
\newblock \bibinfo{booktitle}{\emph{{Sparse sinkhorn attention}}}.
\newblock \bibinfo{type}{{T}echnical {R}eport}.
\newblock
\showISSN{23318422}
\showeprint[arxiv]{2002.11296}


\bibitem[\protect\citeauthoryear{Tay, Juan, Bahri, Zhao, Metzler, and
  Zheng}{Tay et~al\mbox{.}}{2020b}]%
        {Tay}
\bibfield{author}{\bibinfo{person}{Yi Tay}, \bibinfo{person}{Da~Cheng Juan},
  \bibinfo{person}{Dara Bahri}, \bibinfo{person}{Zhe Zhao},
  \bibinfo{person}{Donald Metzler}, {and} \bibinfo{person}{Che Zheng}.}
  \bibinfo{year}{2020}\natexlab{b}.
\newblock \bibinfo{booktitle}{\emph{{SYNTHESIZER: Rethinking Self-Attention in
  Transformer Models}}}.
\newblock \bibinfo{type}{{T}echnical {R}eport}.
\newblock
\showISSN{23318422}
\showeprint[arxiv]{2005.00743}


\bibitem[\protect\citeauthoryear{Vaswani, Shazeer, Parmar, Uszkoreit, Jones,
  Gomez, Kaiser, and Polosukhin}{Vaswani et~al\mbox{.}}{2017}]%
        {Vaswani2017_attentionAllYouNeed}
\bibfield{author}{\bibinfo{person}{Ashish Vaswani}, \bibinfo{person}{Noam
  Shazeer}, \bibinfo{person}{Niki Parmar}, \bibinfo{person}{Jakob Uszkoreit},
  \bibinfo{person}{Llion Jones}, \bibinfo{person}{Aidan~N. Gomez},
  \bibinfo{person}{{\L}ukasz Kaiser}, {and} \bibinfo{person}{Illia
  Polosukhin}.} \bibinfo{year}{2017}\natexlab{}.
\newblock \showarticletitle{{Attention is all you need}}. In
  \bibinfo{booktitle}{\emph{Advances in Neural Information Processing
  Systems}}, Vol.~\bibinfo{volume}{2017-December}. \bibinfo{publisher}{Neural
  information processing systems foundation}, \bibinfo{pages}{5999--6009}.
\newblock
\showISSN{10495258}
\showeprint[arxiv]{1706.03762}
\urldef\tempurl%
\url{https://arxiv.org/abs/1706.03762v5}
\showURL{%
\tempurl}


\bibitem[\protect\citeauthoryear{Verma}{Verma}{2021}]%
        {verma2021beyond}
\bibfield{author}{\bibinfo{person}{Madhusudan Verma}.}
  \bibinfo{year}{2021}\natexlab{}.
\newblock \showarticletitle{Beyond Nystr$\backslash$" omformer--Approximation
  of self-attention by Spectral Shifting}.
\newblock \bibinfo{journal}{\emph{arXiv preprint arXiv:2103.05638}}
  (\bibinfo{year}{2021}).
\newblock


\bibitem[\protect\citeauthoryear{Wang, Li, and Lazebnik}{Wang
  et~al\mbox{.}}{2016}]%
        {Wang2015_Learning_Deep_Structure-Preserving}
\bibfield{author}{\bibinfo{person}{Liwei Wang}, \bibinfo{person}{Yin Li}, {and}
  \bibinfo{person}{Svetlana Lazebnik}.} \bibinfo{year}{2016}\natexlab{}.
\newblock \showarticletitle{{Learning deep structure-preserving image-text
  embeddings}}.
\newblock \bibinfo{journal}{\emph{Proceedings of the IEEE Computer Society
  Conference on Computer Vision and Pattern Recognition}}
  \bibinfo{volume}{2016-December} (\bibinfo{date}{nov} \bibinfo{year}{2016}),
  \bibinfo{pages}{5005--5013}.
\newblock
\showISBNx{9781467388504}
\showISSN{10636919}
\urldef\tempurl%
\url{https://doi.org/10.1109/CVPR.2016.541}
\showDOI{\tempurl}
\showeprint[arxiv]{1511.06078}


\bibitem[\protect\citeauthoryear{Wang, Li, Khabsa, Fang, and Ma}{Wang
  et~al\mbox{.}}{2020a}]%
        {Wang2020a_LinFormer}
\bibfield{author}{\bibinfo{person}{Sinong Wang}, \bibinfo{person}{Belinda~Z.
  Li}, \bibinfo{person}{Madian Khabsa}, \bibinfo{person}{Han Fang}, {and}
  \bibinfo{person}{Hao Ma}.} \bibinfo{year}{2020}\natexlab{a}.
\newblock \showarticletitle{{Linformer: Self-Attention with Linear
  Complexity}}.
\newblock \bibinfo{journal}{\emph{arXiv}} (\bibinfo{date}{jun}
  \bibinfo{year}{2020}).
\newblock
\showISSN{23318422}
\showeprint[arxiv]{2006.04768}
\urldef\tempurl%
\url{http://arxiv.org/abs/2006.04768}
\showURL{%
\tempurl}


\bibitem[\protect\citeauthoryear{Wang, Wang, Yao, Shan, and Chen}{Wang
  et~al\mbox{.}}{2020b}]%
        {Wang2019a_Cross-modal_Scene_Graph_Matching}
\bibfield{author}{\bibinfo{person}{Sijin Wang}, \bibinfo{person}{Ruiping Wang},
  \bibinfo{person}{Ziwei Yao}, \bibinfo{person}{Shiguang Shan}, {and}
  \bibinfo{person}{Xilin Chen}.} \bibinfo{year}{2020}\natexlab{b}.
\newblock \showarticletitle{{Cross-modal scene graph matching for
  relationship-aware image-text retrieval}}.
\newblock \bibinfo{journal}{\emph{Proceedings - 2020 IEEE Winter Conference on
  Applications of Computer Vision, WACV 2020}} (\bibinfo{date}{oct}
  \bibinfo{year}{2020}), \bibinfo{pages}{1497--1506}.
\newblock
\showISBNx{9781728165530}
\urldef\tempurl%
\url{https://doi.org/10.1109/WACV45572.2020.9093614}
\showDOI{\tempurl}
\showeprint[arxiv]{1910.05134}


\bibitem[\protect\citeauthoryear{Wang, Yang, Qian, Ma, Lu, Li, and Fan}{Wang
  et~al\mbox{.}}{2019}]%
        {Wang2019_Position_Focused_Attention}
\bibfield{author}{\bibinfo{person}{Yaxiong Wang}, \bibinfo{person}{Hao Yang},
  \bibinfo{person}{Xueming Qian}, \bibinfo{person}{Lin Ma},
  \bibinfo{person}{Jing Lu}, \bibinfo{person}{Biao Li}, {and}
  \bibinfo{person}{Xin Fan}.} \bibinfo{year}{2019}\natexlab{}.
\newblock \showarticletitle{{Position focused attention network for image-text
  matching}}.
\newblock \bibinfo{journal}{\emph{IJCAI International Joint Conference on
  Artificial Intelligence}}  \bibinfo{volume}{2019-August} (\bibinfo{date}{jul}
  \bibinfo{year}{2019}), \bibinfo{pages}{3792--3798}.
\newblock
\showISBNx{9780999241141}
\showISSN{10450823}
\urldef\tempurl%
\url{https://doi.org/10.24963/ijcai.2019/526}
\showDOI{\tempurl}
\showeprint[arxiv]{1907.09748}


\bibitem[\protect\citeauthoryear{Wei, Zhang, Li, Zhang, and Wu}{Wei
  et~al\mbox{.}}{2020}]%
        {Weia_Multi-Modality_Cross_Attention}
\bibfield{author}{\bibinfo{person}{Xi Wei}, \bibinfo{person}{Tianzhu Zhang},
  \bibinfo{person}{Yan Li}, \bibinfo{person}{Yongdong Zhang}, {and}
  \bibinfo{person}{Feng Wu}.} \bibinfo{year}{2020}\natexlab{}.
\newblock \bibinfo{booktitle}{\emph{{Multi-modality cross attention network for
  image and sentence matching}}}.
\newblock \bibinfo{type}{{T}echnical {R}eport}. \bibinfo{pages}{10938--10947}
  pages.
\newblock
\showISSN{10636919}
\urldef\tempurl%
\url{https://doi.org/10.1109/CVPR42600.2020.01095}
\showDOI{\tempurl}


\bibitem[\protect\citeauthoryear{Wu, Xu, Dai, Wan, Zhang, Tomizuka, Keutzer,
  and Vajda}{Wu et~al\mbox{.}}{2020}]%
        {Wu2020_VisualTransformer}
\bibfield{author}{\bibinfo{person}{Bichen Wu}, \bibinfo{person}{Chenfeng Xu},
  \bibinfo{person}{Xiaoliang Dai}, \bibinfo{person}{Alvin Wan},
  \bibinfo{person}{Peizhao Zhang}, \bibinfo{person}{Masayoshi Tomizuka},
  \bibinfo{person}{Kurt Keutzer}, {and} \bibinfo{person}{Peter Vajda}.}
  \bibinfo{year}{2020}\natexlab{}.
\newblock \showarticletitle{{Visual Transformers: Token-based Image
  Representation and Processing for Computer Vision}}.
\newblock \bibinfo{journal}{\emph{arXiv}} (\bibinfo{date}{jun}
  \bibinfo{year}{2020}).
\newblock
\showISSN{23318422}
\showeprint[arxiv]{2006.03677}
\urldef\tempurl%
\url{http://arxiv.org/abs/2006.03677}
\showURL{%
\tempurl}


\bibitem[\protect\citeauthoryear{Wu, Liu, and Liu}{Wu et~al\mbox{.}}{2021}]%
        {Wu2021_CentroidTransformer}
\bibfield{author}{\bibinfo{person}{Lemeng Wu}, \bibinfo{person}{Xingchao Liu},
  {and} \bibinfo{person}{Qiang Liu}.} \bibinfo{year}{2021}\natexlab{}.
\newblock \showarticletitle{{Centroid Transformers: Learning to Abstract with
  Attention}}.
\newblock  (\bibinfo{date}{feb} \bibinfo{year}{2021}).
\newblock
\showeprint[arxiv]{2102.08606}
\urldef\tempurl%
\url{http://arxiv.org/abs/2102.08606}
\showURL{%
\tempurl}


\bibitem[\protect\citeauthoryear{Xu, Ba, Kiros, Cho, Courville, Salakhutdinov,
  Zemel, and Bengio}{Xu et~al\mbox{.}}{2015}]%
        {Xu2015_Show_attend_and_tell}
\bibfield{author}{\bibinfo{person}{Kelvin Xu}, \bibinfo{person}{Jimmy~Lei Ba},
  \bibinfo{person}{Ryan Kiros}, \bibinfo{person}{Kyunghyun Cho},
  \bibinfo{person}{Aaron Courville}, \bibinfo{person}{Ruslan Salakhutdinov},
  \bibinfo{person}{Richard~S. Zemel}, {and} \bibinfo{person}{Yoshua Bengio}.}
  \bibinfo{year}{2015}\natexlab{}.
\newblock \showarticletitle{{Show, attend and tell: Neural image caption
  generation with visual attention}}. In \bibinfo{booktitle}{\emph{32nd
  International Conference on Machine Learning, ICML 2015}},
  Vol.~\bibinfo{volume}{3}. \bibinfo{publisher}{International Machine Learning
  Society (IMLS)}, \bibinfo{pages}{2048--2057}.
\newblock
\showISBNx{9781510810587}
\showeprint[arxiv]{1502.03044}
\urldef\tempurl%
\url{https://arxiv.org/abs/1502.03044v3}
\showURL{%
\tempurl}


\bibitem[\protect\citeauthoryear{Xu, Li, Yan, Deng, and Liu}{Xu
  et~al\mbox{.}}{2019}]%
        {DBLP:conf/ijcai/XuLYDL19}
\bibfield{author}{\bibinfo{person}{Ruiqing Xu}, \bibinfo{person}{Chao Li},
  \bibinfo{person}{Junchi Yan}, \bibinfo{person}{Cheng Deng}, {and}
  \bibinfo{person}{Xianglong Liu}.} \bibinfo{year}{2019}\natexlab{}.
\newblock \showarticletitle{Graph Convolutional Network Hashing for Cross-Modal
  Retrieval}. In \bibinfo{booktitle}{\emph{Proceedings of the Twenty-Eighth
  International Joint Conference on Artificial Intelligence, {IJCAI} 2019,
  Macao, China, August 10-16, 2019}}, \bibfield{editor}{\bibinfo{person}{Sarit
  Kraus}} (Ed.). \bibinfo{publisher}{ijcai.org}, \bibinfo{pages}{982--988}.
\newblock
\urldef\tempurl%
\url{https://doi.org/10.24963/ijcai.2019/138}
\showDOI{\tempurl}


\bibitem[\protect\citeauthoryear{Yao, Pan, Li, and Mei}{Yao
  et~al\mbox{.}}{2018}]%
        {Yao_2018_ECCV}
\bibfield{author}{\bibinfo{person}{Ting Yao}, \bibinfo{person}{Yingwei Pan},
  \bibinfo{person}{Yehao Li}, {and} \bibinfo{person}{Tao Mei}.}
  \bibinfo{year}{2018}\natexlab{}.
\newblock \showarticletitle{Exploring Visual Relationship for Image
  Captioning}. In \bibinfo{booktitle}{\emph{Proceedings of the European
  Conference on Computer Vision (ECCV)}}.
\newblock


\bibitem[\protect\citeauthoryear{Young, Lai, Hodosh, and Hockenmaier}{Young
  et~al\mbox{.}}{2014}]%
        {DBLP:journals/tacl/YoungLHH14}
\bibfield{author}{\bibinfo{person}{Peter Young}, \bibinfo{person}{Alice Lai},
  \bibinfo{person}{Micah Hodosh}, {and} \bibinfo{person}{Julia Hockenmaier}.}
  \bibinfo{year}{2014}\natexlab{}.
\newblock \showarticletitle{From image descriptions to visual denotations: New
  similarity metrics for semantic inference over event descriptions}.
\newblock \bibinfo{journal}{\emph{Trans. Assoc. Comput. Linguistics}}
  \bibinfo{volume}{2} (\bibinfo{year}{2014}), \bibinfo{pages}{67--78}.
\newblock
\urldef\tempurl%
\url{https://tacl2013.cs.columbia.edu/ojs/index.php/tacl/article/view/229}
\showURL{%
\tempurl}


\bibitem[\protect\citeauthoryear{Zeng, Zhou, Srivastava, Kannan, and
  Prasanna}{Zeng et~al\mbox{.}}{[n.d.]}]%
        {ZENG2021166}
\bibfield{author}{\bibinfo{person}{Hanqing Zeng}, \bibinfo{person}{Hongkuan
  Zhou}, \bibinfo{person}{Ajitesh Srivastava}, \bibinfo{person}{Rajgopal
  Kannan}, {and} \bibinfo{person}{Viktor Prasanna}.}
  \bibinfo{year}{[n.d.]}\natexlab{}.
\newblock \showarticletitle{Accurate, efficient and scalable training of Graph
  Neural Networks}.
\newblock \bibinfo{journal}{\emph{J. Parallel and Distrib. Comput.}}
  (\bibinfo{year}{[n.\,d.]}).
\newblock


\bibitem[\protect\citeauthoryear{Zhang, Lei, Zhang, and Li}{Zhang
  et~al\mbox{.}}{[n.d.]}]%
        {Zhang_contextAware}
\bibfield{author}{\bibinfo{person}{Qi Zhang}, \bibinfo{person}{Zhen Lei},
  \bibinfo{person}{Zhaoxiang Zhang}, {and} \bibinfo{person}{Stan~Z Li}.}
  \bibinfo{year}{[n.d.]}\natexlab{}.
\newblock \bibinfo{booktitle}{\emph{{Context-Aware Attention Network for
  Image-Text Retrieval}}}.
\newblock \bibinfo{type}{{T}echnical {R}eport}.
\newblock


\bibitem[\protect\citeauthoryear{Zhao, Wu, and Luo}{Zhao et~al\mbox{.}}{2021}]%
        {9292444}
\bibfield{author}{\bibinfo{person}{Wentian Zhao}, \bibinfo{person}{Xinxiao Wu},
  {and} \bibinfo{person}{Jiebo Luo}.} \bibinfo{year}{2021}\natexlab{}.
\newblock \showarticletitle{Cross-Domain Image Captioning via Cross-Modal
  Retrieval and Model Adaptation}.
\newblock \bibinfo{journal}{\emph{IEEE Transactions on Image Processing}}
  \bibinfo{volume}{30} (\bibinfo{year}{2021}), \bibinfo{pages}{1180--1192}.
\newblock
\urldef\tempurl%
\url{https://doi.org/10.1109/TIP.2020.3042086}
\showDOI{\tempurl}


\bibitem[\protect\citeauthoryear{Zheng, Zheng, Garrett, Yang, and Shen}{Zheng
  et~al\mbox{.}}{2017}]%
        {Zheng2017_Dual-Path_Convolutional}
\bibfield{author}{\bibinfo{person}{Zhedong Zheng}, \bibinfo{person}{Liang
  Zheng}, \bibinfo{person}{Michael Garrett}, \bibinfo{person}{Yi Yang}, {and}
  \bibinfo{person}{Yi~Dong Shen}.} \bibinfo{year}{2017}\natexlab{}.
\newblock \showarticletitle{{Dual-path convolutional image-text embeddings with
  instance loss}}.
\newblock \bibinfo{journal}{\emph{arXiv}} (\bibinfo{date}{nov}
  \bibinfo{year}{2017}).
\newblock
\showISSN{23318422}
\showeprint[arxiv]{1711.05535}
\urldef\tempurl%
\url{http://arxiv.org/abs/1711.05535}
\showURL{%
\tempurl}


\end{thebibliography}

\end{document}